\documentclass[runningheads]{llncs}

\usepackage{eccv}

\usepackage[dvipsnames]{xcolor}

\usepackage{array,multirow,graphicx}
\usepackage{float}
\usepackage{amsmath}
\usepackage{mathtools}
\usepackage{bbm}
\usepackage{hhline}
\usepackage{boldline}
\usepackage{tabularx}
\usepackage{float}
\usepackage{comment}
\usepackage{color}
\usepackage{amssymb}
\usepackage{booktabs}
\usepackage{multicol}
\usepackage{microtype}      
\usepackage{xcolor}         
\usepackage{algorithm}
\usepackage{algpseudocode}
\usepackage{pifont}
\usepackage{soul} 
\usepackage{subcaption}
\usepackage{wrapfig}
\usepackage{lipsum}

\usepackage{eccvabbrv}

\usepackage{graphicx}
\usepackage{booktabs}
\usepackage{xcolor,colortbl}
\usepackage{nicematrix}
\usepackage[accsupp]{axessibility}  

\usepackage{hyperref}

\usepackage{orcidlink}

\newcommand{\Skip}[1]{}

\begin{document}

\title{Mitigating Background Shift in\\ Class-Incremental Semantic Segmentation}

\titlerunning{MBS}

\author{Gilhan Park\orcidlink{0009-0007-4879-3627},
WonJun Moon\orcidlink{0000-0003-2805-0926},
SuBeen Lee\orcidlink{0009-0005-1470-1160},
Tae-Young Kim\orcidlink{0009-0009-6405-7739},
Jae-Pil Heo\thanks{Corresponding author}\orcidlink{0000-0001-9684-7641}}

\authorrunning{G. Park et al.}

\institute{Sungkyunkwan University, South Korea}

\maketitle
\begin{abstract}
Class-Incremental Semantic Segmentation~(CISS) aims to learn new classes without forgetting the old ones, using only the labels of the new classes.
To achieve this, two popular strategies are employed: 1) pseudo-labeling and knowledge distillation to preserve prior knowledge; and 2) background weight transfer, which leverages the broad coverage of background in learning new classes by transferring background weight to the new class classifier.
However, the first strategy heavily relies on the old model in detecting old classes while undetected pixels are regarded as the background, thereby leading to the background shift towards the old classes~(\textit{i.e.}, misclassification of old class as background). 
Additionally, in the case of the second approach, initializing the new class classifier with background knowledge triggers a similar background shift issue, but towards the new classes.
To address these issues, we propose a background-class separation framework for CISS.
To begin with, selective pseudo-labeling and adaptive feature distillation are to distill only trustworthy past knowledge.
On the other hand, we encourage the separation between the background and new classes with a novel orthogonal objective along with label-guided output distillation.
Our state-of-the-art results validate the effectiveness of these proposed methods. Our code is available at: \url{https://github.com/RoadoneP/ECCV2024_MBS}.
\end{abstract}
\section{Introduction}
Semantic segmentation, fundamental for applications such as autonomous driving and medical imaging~\cite{FullyCrfs, Segmenter, robotics, U-net}, is a task to classify pixel-wise semantics within predefined classes on specific datasets. 
However, in practical scenarios, it is often necessary for models to learn additional classes after deployment.
The primary goal of Class-Incremental Semantic Segmentation~(CISS) is to extend the model's capability only with the supervision of newly introduced classes, without forgetting the old ones. 

\begin{figure}[t!]
    \centering
    \includegraphics[width=1\textwidth]{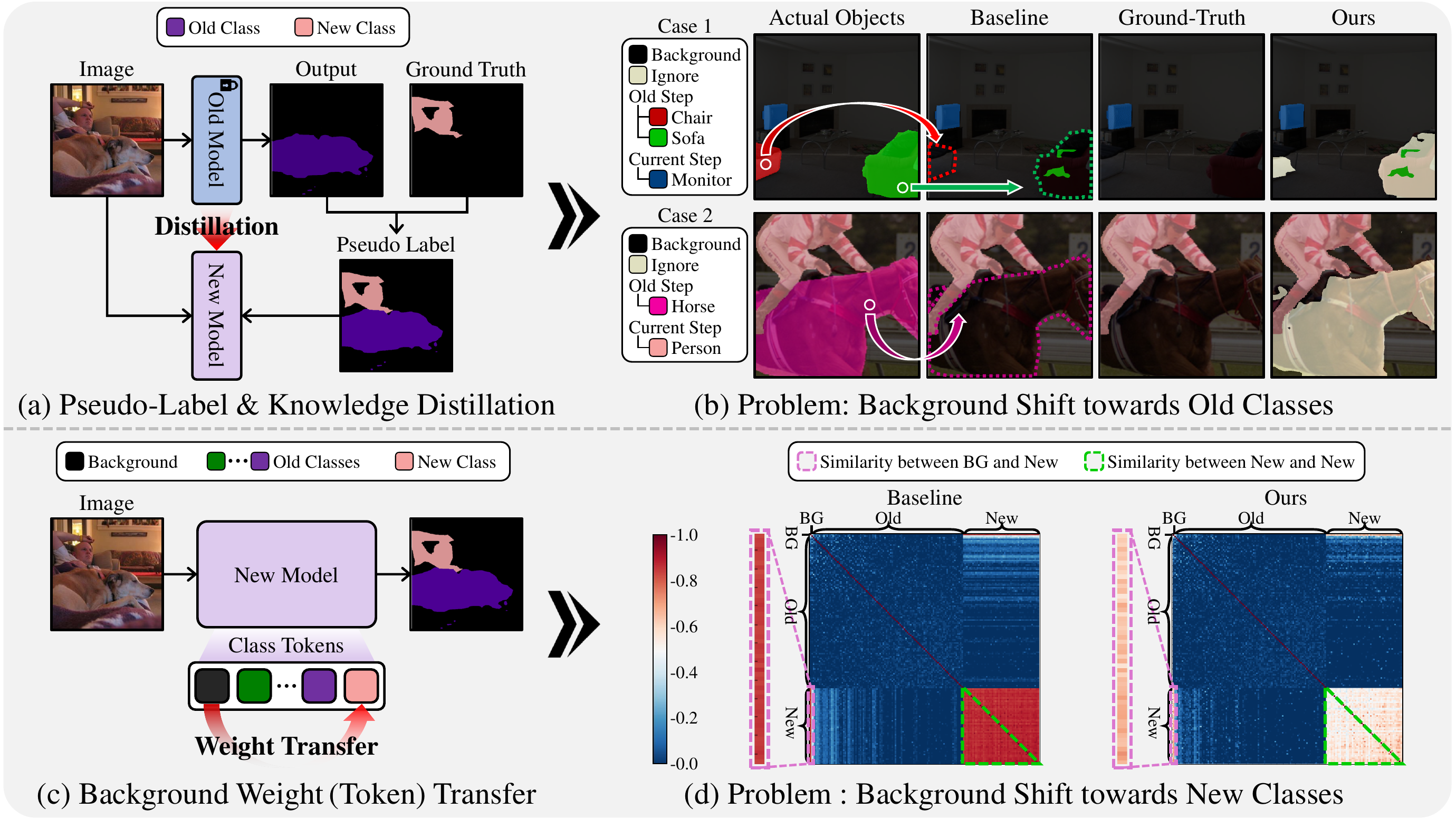}
    \caption{
    (a) Pseudo-labeling is used to learn unlabeled old classes in the image based on the prediction of the old model, while knowledge distillation is to retain the intermediate knowledge of old classes by minimizing the difference between features from the old and new models.
    (b) However, typical models hardly recognize all pixels precisely. Therefore, the ambiguous pixels~(with low prediction confidences from old model) are labeled as the background, \textit{i.e.}, chair and sofa in Case~1 and horse in Case~2, causing the background shift towards old classes~(\textit{i.e.}, misclassification of old classes as background). 
    In contrast, our method alleviates the background shift towards old classes by ignoring these ambiguous pixels.
    (c) On the other hand, background weight transfer leverages the broad category coverage of the background class by initializing the new class token with the background token parameters.
    (d) Despite its advantages, the baseline model faces challenges in clearly distinguishing new class tokens from the background ones~(i.e., background shift towards new classes). Conversely, our method demonstrates improved separation of these classes while preserving the benefits of background weight transfer.
    }
    \label{fig:fig1}
\end{figure}

To achieve this, existing strategies~\cite{PLOP, ssul, RBC, Incrementer} have adopted 1) pseudo-labeling and knowledge distillation to retain old class knowledge, \textit{i.e.}, prevent catastrophic forgetting~\cite{catastrophicforgetting, catastrophicforgetting2}, as described in Fig.~\ref{fig:fig1}~(a).
Furthermore, 2) background weight transfer is another popular technique~\cite{MiB} that duplicates background classifier weight to new class weight to exploit the broad semantic coverage of the background in learning new classes as illustrated in Fig.~\ref{fig:fig1}~(c).
Nevertheless, we point out that these strategies remain vulnerable to the issue of background shift, which complicates the distinction between background and object classes.
Specifically, the first strategy highly relies on the old model and regards ambiguous pixels as the background, triggering the background shift towards the old classes, as described in Fig.~\ref{fig:fig1}~(b).
On the other hand, the second approach triggers the background shift towards new classes since the separation of the new class from the background is a challenging problem, as illustrated in Fig.~\ref{fig:fig1}~(d).

With these motivations, we propose a background-class separation framework for CISS to prevent background shifts towards both old and new classes.
Initially, we propose a selective strategy during the pseudo-labeling process.
To illustrate, whereas the pixels with low prediction confidence from the old model are regarded as the background class for pseudo-labeling, our strategy selectively assigns the background label only if the probability of being the actual object is low.
This approach effectively reduces the background shift towards old classes by preventing the possible old class samples from being treated as the background.
Similarly, adaptive feature distillation shares the same motivation that features for corresponding ambiguous patches are not trustworthy; we use the patch-wise prediction confidence as the weight to calibrate the degree of distillation for each patch.
Although we mitigate the background shift towards old classes with two formerly discussed distillation methods, the background shift towards new classes remains due to the background weight transfer strategy.
To tackle this, we employ label-guided output distillation which enables the decoupling of background and new classes~\cite{LGKD} while benefiting from the background weight transfer strategy.
In addition, we also introduce the novel orthogonal objective between the new classes and background tokens to further mitigate the background shift towards new classes.

To sum up, our contributions are as follows:
\begin{itemize}
    \setlength\itemsep{1pt}
    \item[$\bullet$] We devise a selective pseudo-labeling strategy that excludes ambiguous pixels to prevent the object pixels from being misclassified as the background, mitigating the background shift towards old classes.
    \item[$\bullet$] To further alleviate the background shift towards old classes, we propose an adaptive feature distillation to distill only the reliable representations. 
    \item[$\bullet$] We introduce an orthogonal objective to mitigate the background shift towards new classes while keeping the advantage of it.
    \item[$\bullet$] The benefits of each component and the superiority of our full method are validated with ablation studies and state-of-the-art results on all datasets.
\end{itemize} 
\section{Related Work}

\subsection{Incremental Learning}
Recently, incremental learning has been extensively studied to overcome the catastrophic forgetting~\cite{catastrophicforgetting, catastrophicforgetting2} issue of deep learning models in incremental scenarios. 
Previous works fall into three major categories: 1) regularization-based, 2) architectural-based, and 3) replay-based. 
Regularization-based methods~\cite{EWC, LWM, riemann, csi} aim to devise the regularizer that is to alleviate the forgetting problem by preventing the model parameters from drastic changes. On the other hand, architectural-based methods~\cite{packnet, adaptivenetwork, der, l2g} focus on proposing a new network architecture or developing adaptive designs for new tasks. 
Finally, replay-based methods~\cite{ganreplay, lucir, icarl, gcr} store the past data or utilize the generative model to replay the simplified version of the old task while learning the new tasks.
Among these streams, our proposed method falls into the first category, \textit{i.e.}, regularization-based.
Yet, our goal is to highlight the significance of being discerned when distilling the prior knowledge to alleviate background shift.


\subsection{Class-Incremental Semantic Segmentation}
Semantic segmentation is a per-pixel classification task that aims to classify each pixel in the target image~\cite{FullyCrfs, Segmenter, U-net, HP}. 
Employing popular convolutional networks~\cite{FCN, maskrcnn, DAnet,deeplabv3, Ccnet} and transformer-based architectures~\cite{segformer, Segmenter, maskformer}, we have observed the drastic improvements in semantic segmentation within the past few years.
However, these methods are susceptible to incremental scenarios due to catastrophic forgetting. 
Tackling such vulnerability, the problem of Class-Incremental Semantic Segmentation has been proposed by ILT~\cite{ILT}.
Since then, the need for CISS has been highlighted; MiB~\cite{MiB} utilized unbiased knowledge distillation to address the background shift problem, PLOP~\cite{PLOP} used the multi-scale distillation and pseudo-labeling strategy to maintain past knowledge, and Incrementer~\cite{Incrementer} proposed transformer-based framework and eased the learning process of new classes by distilling only the old-classes-relevant features.
Our work resembles \cite{Incrementer, MiB, PLOP} in that we utilize prior knowledge to prevent the catastrophic forgetting of the old classes.
However, our methods differ in that our goal is to effectively address background shift by separating objects (either of old and new classes) and the background.
\section{Method}
\subsection{Problem Formulation}
Class-Incremental Semantic Segmentation~(CISS) aims for the model to learn new classes while retaining previously learned knowledge, using supervision only for the novel classes. 
Basically, the model for CISS is sequentially trained across multiple timesteps~$(t=1, 2, \dots, T)$, and its stability in preserving old classes is evaluated along with its plasticity in learning new classes at each step. 
At each time step $t$, a set of classes $C^t$ newly appears and is distinct, ensuring that there is no overlap with the class sets from other steps formally denoted as~($C^i \cap C^j = \emptyset ~\text{for all } i \neq j, \text{ where } 1 \leq i, j \leq T$).  
Then, the model learns new classes with data $D^t= \{ (x^t_i \in \mathbb{R}^{H^I \times W^I \times 3}, y^t_i\in\mathbb{R}^{H^I \times W^I} ) \}_{i=1}^{N}$, where $H^I$ and $W^I$ denote the height and width of the image, respectively, and $x^t_i$, $y^t_i$ represent the image, corresponding label map, respectively.
Since only the supervision for novel classes is given at each step, the label map $y^t$ consists of a set of novel classes $c$ in $C^t$ and the background class $c_0$.
Thus, the only available annotation in learning at $t$-th time step is whether each pixel belongs to the novel classes or not. 
In short, the pixels labeled as the background class in step $t$ may contain the pixels that belong to the seen class sets $C^{1:t-1}$ or future class sets $C^{t+1:T}$.
\begin{figure}[t]
    \centering
    \includegraphics[width=1.\textwidth]{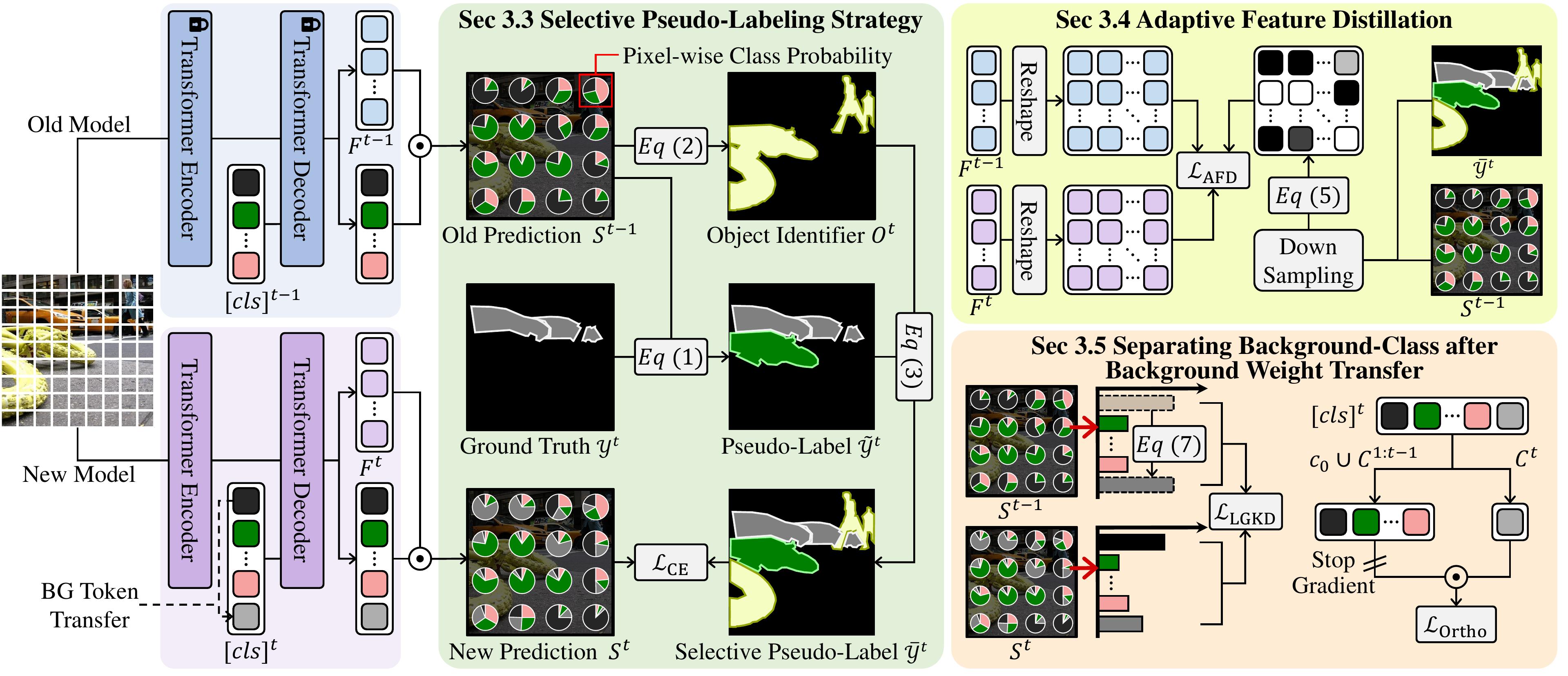}
    \caption{
    Overview of our background-class separation framework for CISS.
        Given a new class `car'~(\textbf{grey}) at step $t$~(\textbf{purple}), a new class token is added to $[cls]^t$ where the token weight is initially duplicated from the background class.
    (\textbf{Blue}) Initially, the image is fed into the old model from the previous step $t-1$ to generate the prediction $S^{t-1}$, (\textbf{green}) which is then used to calculate the object identifier $O^t$ and the pseudo-label $\tilde{y}^t$. 
    Subsequently, these are combined to produce the selective pseudo-label map $\bar{y}^t$ to train the new model at step $t$.
    (\textbf{yellow}) 
    Along with the selective pseudo-labeling, we further calibrate the degree of distilling the old knowledge based on patch-wise reliability, \textit{i.e.}, prediction confidence, through adaptive feature distillation $\mathcal{L}_\text{AFD}$.
    Briefly, the degree for each patch to be distilled is derived from the combination of $S^{t-1}$ and $\bar{y}^t$.
    (\textbf{orange}) For the semantic separation of the new class from the background at step $t$, background probability of old model is distilled into the new class probability of the new model.
    To further support the separation, an orthogonality loss between the new class and the background is implemented.
    }
    \label{fig:main_figure}
\end{figure}

\subsection{Overview}
The overview of our model is illustrated in \cref{fig:main_figure}. 
Employing Segmenter~\cite{Segmenter} with popular techniques for CISS, \textit{i.e.}, Pseudo-Labeling~(PL), Knowledge Distillation~(KD), and the Background token Transfer~(BT)~\cite{MiB}, as the base framework, we propose multiple strategies to address the background shift in CISS.
Despite the recognized their effectiveness of PL, KD, and BT, we claim that the background shift remains a significant unresolved challenge.
First, the general protocol of PL is to regard the ambiguous pixels~(lower than the threshold of PL) in the perspective of the old model as the background class which leads to a background shift towards old classes.
Although KD is another method to distill the old model's knowledge, it forwards all features, including those that do not reflect the actual semantics, since conventional KD does not consider the old model's reliability.
In continual scenarios, this leads to an error propagation problem~(often triggering the background shift due to the large semantic coverage of the background).
Finally, BT is to exploit the large semantic coverage in the background so that future classes are semantically included in the coverage of the background.
Thus, it is shown to ease the initial training process of new classes when the background classifier weights are transferred to the ones of new classes.
Yet, as the initial points of background and new classes are the same, it naturally makes the background shift towards the new classes.
To address these, we propose selective pseudo-labeling~(Sec.~\ref{sec.3.3}), adaptive feature distillation~(Sec.~\ref{sec.3.4}), and orthogonal objective (Sec.~\ref{sec.3.5}) to mitigate the background shift problem caused by aforementioned in CISS.

\subsection{Selective Pseudo-Labeling Strategy}
\label{sec.3.3}
At every continual step in CISS, labels are provided only for the pixels belonging to the new classes. 
In other words, the pixels categorized as background may be part of either the old classes, including the semantic background class, or future classes. 
Because treating them as background could exacerbate catastrophic forgetting due to background shift towards old classes, many previous methods~\cite{MiB, PLOP, Incrementer} have employed PL to address the issue.
Generally, PL is conventional to assign each pixel with the specific object class as the pseudo-label if the prediction confidence of the old model for each pixel surpasses a predefined threshold $\tau$ and otherwise labeled as a background class~\cite{PLOP, RBC, ssul, Incrementer}. 
Therefore, the pseudo-label map $\tilde{y}\in \mathbb{R}^{H^I \times W^I}$~\cite{PLOP} is described below:
\begin{gather}
    \begin{aligned}
        \tilde{y}^t_{h, w}=
        \begin{cases}
            y^t_{h,w} & \text{if }({y}^t_{h,w} \neq c_0)
            \\
            \underset{c \in C^{1:t-1}}{\mathrm{argmax}} S^{t-1}_{h,w,c} & \text{if } (y^t_{h,w} = c_0) \wedge (\exists_{c\in C^{1:t-1}}S^{t-1}_{h,w,c} >\tau )
            \\[-4pt]
            c_0 & \text{otherwise}
        \end{cases}
    \end{aligned}
    \raisetag{15pt}
\end{gather}
where $S^{t-1} \in \mathbb{R}^{H^I \times W^I \times \left(\mid C^{1:t-1} \mid + 1\right)}$ denotes the segmentation prediction of the old model after applying the softmax.

However, these methods highly rely on the old model's capability to generate pseudo-labels for all other unlabeled pixels, and if the predictions of the old model are incorrect, this can result in pixels of the old classes being mislabeled as the background in the new model's training. 
As a result, this aggravates the background shift towards old classes.

To address the above problem, we suggest being selective when distilling the knowledge of the old model via PL. 
This involves defining the object identifier $O$ for each pixel, based on the old model's pixel-wise prediction confidence, to identify pixels distinct from the background class.
We design the object identifier to be 1 when the sum of confidence scores over object classes is bigger than that of the background class. This is because the low confidence for the background is a strong signal that the pixel is highly likely to contain salient objects.
Formally, the object identifier $O^t \in \mathbb{R}^{H^I \times W^I}$ at step $t$ is defined as:
\begin{equation}
    O^t_{h,w}=
    \begin{cases}
        1 & \text{if } {S}^{t-1}_{h,w,c_{0}} < \sum_{c \in C^{1:t-1}}S^{t-1}_{h,w,c}\\
        0 & \text{otherwise}
    \end{cases}
\end{equation}
Then, we calculate the selective pseudo-label map with the object identifier.
we apply the selection strategy to detect background-misclassified pixels as below:
\begin{gather}
\label{eq3}
    \begin{aligned}
        \bar{y}^t_{h, w}=
        \begin{cases}
            \tilde{y}^t_{h,w} & \text{if } (\tilde{y}^t_{h,w}\neq c_0)
            \\
            c_0  & \text{if } (\tilde{y}^t_{h,w}=c_0) \wedge (O^t_{h,w}=0)
            \\
            c_{\text{ignore}} & \text{otherwise }
        \end{cases}
    \end{aligned}
\end{gather}
where $\bar{y}^t \in \mathbb{R}^{H^I \times W^I}$ is the final pseudo-label map for step $t$ and $c_{\text{ignore}}$ indicates whether the pixel is excluded in training.
By detecting the objects and preventing them from being labeled as the background class, we expect to mitigate the background shift. 
Consequently, our cross-entropy loss is expressed as:
\begin{gather}
    \label{eq4}
    \begin{aligned}
        \mathcal{L}_{\text{CE}}= -\frac{1}{H^IW^I}\sum_{h=1}^{H^I}\sum_{w=1}^{W^I} \mathbbm{1}_{\bar{y}^{t}_{h,w} \neq c_{\text{ignore}}}  \text{log}(S^{t}_{h,w,\bar{y}^{t}_{h,w}}),
    \end{aligned}
\end{gather}
where \(\mathbbm{1}_{\alpha} = 1\) if \(\alpha\) is true, and \(0\) otherwise.

\subsection{Adaptive Feature Distillation}
\label{sec.3.4}
Along with the PL, KD is another tool to alleviate catastrophic forgetting~\cite{PodNet, LWM, lucir, PLOP, Incrementer}. 
Despite its effectiveness in previous works, we point out that na\"ive KD methods without consideration of the old model's confidence can induce the background shift towards old classes.

To address the issue, we propose an adaptive feature distillation strategy, which accounts for the old model's extensive background knowledge and adaptively distills trustworthy features.
Given the transformer backbone, let us denote the patch features at step $t$ as ${F^t} \in \mathbb{R}^{H^{F} \times W^{F} \times D}$ where $H^{F}=\frac{H^{I}}{P}$ and $W^{F}=\frac{W^{I}}{P}$.
Here, $P$ indicates the patch size.
Then, whereas the general feature distillation protocol is to teach ${F}^t$ to mimic ${F}^{t-1}$, we aim to identify the unreliable patches that may trigger the error propagation.
To achieve this, we utilize a patch reliability map $M$ that measures the reliability of each patch.
Specifically, we design the reliability map to retain the value of 1 in patches where pseudo-labels are assigned to one of the learned object classes with high confidence, signifying a strong learning signal. 
Conversely, we assign a value of 0 to patches corresponding spatially to $c_{\text{ignore}}$ assigned in \cref{eq3} or new classes; this is due to the old model's lack of knowledge about new classes and uncertainty in ambiguous regions. Additionally, for the patches pseudo-labeled as the background class, we assign a value corresponding to each patch's confidence score.
Formally, the reliability map at step $t$, $M^t \in \mathbb{R}^{H^F \times W^F}$ is defined as:
\begin{equation}
    \label{eq5}
    M^t_{h,w}=
    \begin{cases}
        1 & \text{if } (\hat{y}^t_{h,w} \in C^{1:t-1}) \\ 
        0 & \text{if } (\hat{y}^t_{h,w} \in C^{t}) \vee (\hat{y}^t_{h,w} = c_{\text{ignore}})\\
        \hat{S}^{t-1}_{h,w,c_0} & \text{if } (\hat{y}^t_{h,w} = c_0) \\
    \end{cases}
\end{equation}
where $\hat{y}\in\mathbb{R}^{H^F \times W^F}$ and $\hat{S}\in\mathbb{R}^{H^F \times W^F}$ are the downsampled versions of the selective pseudo-label $\bar{y}$~(defined in \cref{eq3}) and segmentation prediction $S$ to address the spatial mismatch with $M$ at the feature level, and $M_{h,w}$ denotes the reliability score of the patch spatially located in~$\left(h,w\right)$, respectively. 
We follow previous works~\cite{ConSeg, sats, uac} to utilize the interpolation for downsampling.

In short, the old model's prediction for a patch located at~$(h, w)$ is considered to be reliable if the value of $M^{t}_{h,w}$ is close to 1.
Consequently, we employ the reliability map as the weight map in distilling the old model's patch-wise knowledge to prevent the new model from learning possibly incorrect supervision.  
With the reliability-based weighting scheme, our adaptive feature distillation loss is expressed as:
\begin{equation}
    \label{eq6}
    \mathcal{L}_{\text{AFD}} = \frac{1}{H^{F}W^{F}} \sum_{h=1}^{H^{F}}\sum_{w=1}^{W^{F}} M^{t}_{h,w}\left\|
F^t_{h, w} - {F}^{t-1}_{h, w} \right\|^2.
\end{equation}
Note that adaptive feature distillation is implemented after the decoding layers following existing works~\cite{MiB, Incrementer}.

\subsection{Separating Background-Class after Background Weight Transfer}
\label{sec.3.5}
The background pixels predicted by the old model may include the novel classes at the current step.
In this regard, MiB~\cite{MiB} eased the process of learning new classes by transferring the knowledge of the background to the new classes. 
Yet, this triggers the background shift towards the new classes, resulting in high correlations between the background and new classes as shown in Fig.~\ref{fig:fig1}~(d).

To tackle this, we adopt a label-guided knowledge distillation~(LGKD)~\cite{LGKD}, distilling the old model's refined logit $\bar{S}^{t-1}$ in which the background logit is transferred to the ones of new classes~(when the pixel belongs to the new class). 
The refined logits $\bar{S}^{t-1}_{h,w,c}$ and the loss for LGKD are expressed as follows:
\begin{gather}
    \label{eq7}
    \begin{aligned}
        \bar{S}^{t-1}_{h,w,c}=
        \begin{cases}
            0 & \text{if } \{(c=c_0) \land (\bar{y}^t_{h,w} \in C^t)\} \vee \{(c\in C^t) \land (\bar{y}^t_{h,w} \neq c)\}
            \\
            S^{t-1}_{h,w,c_0} & \text{if } (c \in C^t) \land (\bar{y}^t_{h,w}=c)
            \\
            S^{t-1}_{h,w,c} & \text{otherwise,}
        \end{cases}
    \end{aligned}
\end{gather}
\begin{gather}
    \label{eq8}
    \begin{aligned}
        \mathcal{L}_{\text{LGKD}}= -\frac{1}{HW} \sum_{h=1}^{H} \sum_{w=1}^W \bar{S}^{t-1}_{h,w} \log S^{t}_{h,w}.
    \end{aligned}
\end{gather}

However, we argue that applying only the distillation objective does not sufficiently separate between the background and new classes~(discussed in \cref{ablation_similarity}).
Therefore, we propose an orthogonal loss to the correlations between the classifier weights of the new class and other classes with a novel view to directly train to separate the new classes from the old ones including the background class.
Formally, the process of applying orthogonality in the classifier weights is expressed as:
\begin{equation}
    \label{eq9}
    \small
    \mathcal{L}_{\text{Ortho}}=\frac{1}{\mid C^{t} \mid \mid C^{1:t} \mid }\sum_{c_i\in {C^t}}\sum_{c_j\in(C^{1:t}\cup\{c_0\})}  {\mathbbm{1}_{i\neq j} \lvert[cls]^t_{c_{i}}\odot \text{sg}([cls]^t_{c_{j}})} \rvert, 
\end{equation}
where $[cls]_{c_{i}}$ denotes the $i$-th class weight and $\text{sg}(\cdot)$ stands for stop gradient operation~\cite{simsiam, byol}.
We use the stop gradient operation on the old class weights to prevent the significant modification of the previously learned embedding space.
Consequently, the background shift towards the new classes can be alleviated as the new class logit learns to deviate from the background class weight.
For the overall loss formulation, we combine $\mathcal{L}_{\text{Ortho}}$ with $\mathcal{L}_{\text{LGKD}}$ as they are compatible with one another in that they share the same goal of separating the new class from the old ones but with different strategies.
We express it as follows:
\begin{gather}
    \label{eq10}
    \begin{aligned}
        \mathcal{L}_{\text{Sep}} = \lambda_{\text{LGKD}}\mathcal{L}_{\text{LGKD}} + \lambda_{\text{Ortho}}\mathcal{L}_{\text{Ortho}},
    \end{aligned}
\end{gather}
where $\lambda_{\text{LGKD}}$ and $\lambda_{\text{Ortho}}$ are coefficients for each objective.

\subsection{Final Objective}
Our proposed methods employ cross-entropy with selective pseudo-labeling $\mathcal{L}_{\text{CE}}$ as described in \cref{eq4}. Also, we use the adaptive feature distillation loss $\mathcal{L}_{\text{AFD}}$ as described in \cref{eq6} and loss objective for background-class separation \cref{eq10}.
To sum it up, our overall objective $\mathcal{L}$ is expressed as:
\begin{gather}
    \begin{aligned}
        \mathcal{L} = \mathcal{L}_{\text{CE}} + \mathcal{L}_{\text{AFD}} + \mathcal{L}_{\text{Sep}}.
    \end{aligned}
\end{gather}
\section{Experiment}
\subsection{Experimental Details}
\subsubsection{Datasets.} Following the experimental setting of ~\cite{Incrementer}, we evaluate our method on two public datasets: Pascal VOC~\cite{pascalvoc} and ADE20k~\cite{ade20k}. 
Pascal VOC contains 10,582 fully annotated images for training and 1,449 images for testing, over 20 foreground object classes. 
ADE20k has 20,210 training images and 2,000 testing images in 150 classes. 
\subsubsection{Experimental Protocols.}
We evaluate the performance of our method under two different CISS settings following MiB~\cite{MiB}: $\textit{Disjoint}$ and $\textit{Overlapped}$. 
In both settings, labels are provided only for the class set $C^t$ that newly appears in the current step $t$. 
However, data $D^t$ is composed of pixels of classes belonging to the old or new class sets~$\left(C^{1:t-1} \cup C^t \right)$ in the \textit{disjoint} setting.
In the \textit{overlapping} setting, pixels from both sets appear, and even those from future class sets~$\left(C^{1:t-1} \cup C^t \cup C^{t+1:T}\right)$.
Among them, the overlapped setting is usually considered more challenging and more realistic in the CISS scenario. 
Additionally, we follow previous~\cite{MiB, Incrementer, RBC} to organize the class compositions with the number of continual steps.
For example, the scenario named 15-1~(6 steps) indicates that we initially train the model on 15 classes and continually learn 1 additional class at every continual step.
For evaluation metrics, we use the mean Intersection over Union~(mIoU). 
Specifically, we compute the mIoU for the initial class set $C^1$, incremented sets $C^{2:T}$, and all class sets $C^{1:T}$~(overall), respectively.
Briefly, each mIoU score in order can be seen as a measure to evaluate the robustness to catastrophic forgetting, plasticity to new classes, and the balance between the two. 
The experimental results of our method are compared with those of previous CISS methods, such as EWC~\cite{EWC}, ILT~\cite{ILT}, MiB~\cite{MiB}, SDR~\cite{SDR}, PLOP~\cite{PLOP}, RECALL~\cite{RECALL}, REMIND~\cite{REMIND}, RCIL~\cite{RC}, SPPA~\cite{SPPA}, RBC~\cite{RBC}, and INC~\cite{Incrementer}. `Joint' denotes the results when all classes are trained at once~(oracle scenario).
\begingroup
\setlength{\tabcolsep}{1pt} 
\renewcommand{\arraystretch}{1.1} 
\begin{table}[t]
    \caption{Performance comparison on Pascal VOC under various scenarios. CNN and Transformer column indicates the type of the backbone network. ${\dagger}$ indicates the reproduced results from \cite{PLOP, RBC, Incrementer}}
    \label{table_pascalvoc}
    {\scriptsize
    \begin{tabular}{l|ccc|ccc|ccc|ccc|ccc|ccc}
    \hlineB{2.5}
    \multicolumn{1}{c|}{} & \multicolumn{6}{c|}{19-1~(2 steps)} & \multicolumn{6}{c|}{15-5~(2 steps)} & \multicolumn{6}{c}{15-1~(6 steps)} \\ \cline{2-19}
    Method & \multicolumn{3}{c|}{Disjoint} & \multicolumn{3}{c|}{Overlapped} & \multicolumn{3}{c|}{Disjoint} & \multicolumn{3}{c|}{Overlapped} & \multicolumn{3}{c|}{Disjoint} & \multicolumn{3}{c}{Overlapped} \\
    & \multicolumn{1}{c}{1-19} & \multicolumn{1}{c}{20} & \multicolumn{1}{c|}{All} & \multicolumn{1}{c}{1-19} & \multicolumn{1}{c}{20} & \multicolumn{1}{c|}{All} & \multicolumn{1}{c}{1-15} & \multicolumn{1}{c}{16-20} & \multicolumn{1}{c|}{All} & \multicolumn{1}{c}{1-15} & \multicolumn{1}{c}{16-20} & \multicolumn{1}{c|}{All} & \multicolumn{1}{c}{1-15} & \multicolumn{1}{c}{16-20} & \multicolumn{1}{c|}{All} & \multicolumn{1}{c}{1-15} & \multicolumn{1}{c}{16-20} & \multicolumn{1}{c}{All} \\ \hline 
    \multicolumn{19}{c}{\textbf{CNN-based Methods}} \\ \hline
    EWC${\dagger}$ & 23.2 & 16.0 & 22.9 & 26.9 & 14.0 & 26.3 & 26.7 & 37.7 & 29.4 & 24.3 & 35.5 & 27.1 & 0.3 & 4.3 & 1.3 & 0.3 & 4.3 & 1.3 \\
    ILT${\dagger}$ & 69.1 & 16.4 & 66.4 & 67.8 & 10.9 & 65.1 & 63.2 & 39.5 & 57.3 & 67.1 & 39.2 & 60.5 & 3.7 & 5.7 & 4.2 & 8.8 & 8.0 & 8.6 \\
    MiB${\dagger}$ & 69.6 & 25.6 & 67.4 & 71.4 & 23.6 & 69.2 & 71.8 & 43.3 & 64.7 & 76.4 & 50.0 & 70.1 & 46.2 & 12.9 & 37.9 & 34.2 & 13.5 & 29.3 \\
    SDR${\dagger}$ & 69.9 & 37.3 & 68.4 & 69.1 & 32.6 & 67.4 & 73.5 & 47.3 & 67.2 & 75.4 & 52.6 & 69.9 & 59.2 & 12.9 & 48.1 & 44.7 & 21.8 & 39.2 \\
    PLOP${\dagger}$ & 75.4 & 38.9 & 73.6 & 75.4 & 37.4 & 73.5 & 71.0 & 42.8 & 64.3 & 75.7 & 51.7 & 70.1 & 57.9 & 13.7 & 46.5 & 65.1 & 21.1 & 54.6 \\
    RECALL & 65.2 & 50.1 & 65.8 & 67.9 & 53.5 & 68.4 & 66.3 & 49.8 & 63.5 & 66.6 & 50.9 & 64.0 & 66.0 & 44.9 & 62.1 & 65.7 & 47.8 & 62.7 \\
    REMIND & \multicolumn{1}{c}{-} & \multicolumn{1}{c}{-} & \multicolumn{1}{c|}{-} & 76.5 & 32.3 & 74.4 & \multicolumn{1}{c}{-} & \multicolumn{1}{c}{-} & \multicolumn{1}{c|}{-} & 76.1 & 50.7 & 70.1 & \multicolumn{1}{c}{-} & \multicolumn{1}{c}{-} & \multicolumn{1}{c|}{-} & 68.3 & 27.2 & 58.5 \\
    RCIL & \multicolumn{1}{c}{-} & \multicolumn{1}{c}{-} & \multicolumn{1}{c|}{-} & \multicolumn{1}{c}{-} & \multicolumn{1}{c}{-} & \multicolumn{1}{c|}{-} & 75.0 & 42.8 & 67.3 & 78.8 & 52.0 & 72.4 & 66.1 & 18.2 & 54.7 & 70.6 & 23.7 & 59.4 \\
    SPPA & 75.5 & 38.0 & 73.7 & 76.5 & 36.2 & 74.6 & 75.3 & 48.7 & 69.0 & 78.1 & 52.9 & 72.1 & 59.6 & 15.6 & 49.1 & 66.2 & 23.3 & 56.0 \\
    RBC${\dagger}$ & 76.4 & 45.8 & 75.0 & 77.3 & 55.6 & 76.2 & 75.1 & 49.7 & 69.9 & 76.6 & 52.8 & 70.9 & 61.7 & 19.5 & 51.6 & 69.5 & 38.4 & 62.1 \\ \cline{1-1}  \cline{1-19}
    Joint & 77.4 & 78.0 & 77.4 & 77.4 & 78.0 & 77.4 & 79.1 & 72.6 & 77.4 & 79.1 & 72.6 & 77.4 & 79.1 & 72.6 & 77.4 & 79.1 & 72.6 & 77.4 \\ \hline
    \multicolumn{19}{c}{\textbf{Transformer-based Methods}} \\ \hline
    MiB${\dagger}$ & 80.6 & 45.2 & 79.6 & 79.9 & 47.7 & 79.1 & 75.0 & 59.9 & 72.3 & 78.6 & 63.1 & 75.6 & 66.7 & 26.3 & 58.3 & 72.6 & 23.1 & 61.7 \\
    RBC${\dagger}$ & 80.9 & 42.1 & 79.7 & 80.2 & 38.8 & 79.0 & 77.7 & 59.1 & 74.0 & 78.9 & 62.0 & 75.5 & 69.0 & 28.4 & 60.5 & 75.9 & 40.2 & 68.2 \\
    INC${\dagger}$ & 82.4 & 64.2 & 82.2 & 82.5 & 61.0 & 82.1 & 81.6 & 62.2 & 77.6 & 82.5 & 69.2 & 79.9 & 81.4 & 57.1 & 76.2 & 79.6 & 59.6 & 75.6 \\
    \rowcolor{gray!30}
    \textbf{Ours} & \textls[-50]{\textbf{82.8}} & \textls[-50]{\textbf{69.3}} & \textls[-50]{\textbf{82.8}} & \textls[-50]{\textbf{83.3}} & \textls[-50]{\textbf{72.0}} & \textls[-50]{\textbf{83.3}} & \textls[-50]{\textbf{81.9}} & \textls[-50]{\textbf{67.2}} & \textls[-50]{\textbf{79.0}} & \textls[-50]{\textbf{84.1}} & \textls[-50]{\textbf{76.0}} & \textls[-50]{\textbf{82.6}} & \textls[-50]{\textbf{81.5}} & \textls[-50]{\textbf{64.7}} & \textls[-50]{\textbf{78.1}} & \textls[-50]{\textbf{82.6}} & \textls[-50]{\textbf{72.2}} & \textls[-50]{\textbf{80.6}} \\ \cline{1-1}  \cline{1-19}
    Joint & 83.9 & 81.6 & 83.8 & 83.9 & 81.6 & 83.8 & 84.8 & 80.7 & 83.8 & 84.8 & 80.7 & 83.8 & 84.8 & 80.7 & 83.8 & 84.8 & 80.7 & 83.8 \\
    \hlineB{2.5}
\end{tabular}
}
\end{table}
\endgroup

\subsubsection{Implementation Details.}
We utilize the ImageNet-pretrained~\cite{imagenet} vision transformer VIT-B/16~\cite{vit} as our encoder and two transformer layers as the decoder~\cite{Segmenter} which processes cropped input image of resolution 512 $\times$ 512.
Our initial learning rate is set to $1\times 10^{-3}$ with SGD~\cite{SGD} optimizer.
Momentum and the weight decay parameters for SGD are set to 0.9 and $1\times 10^{-5}$, respectively, and the pseudo-labeling threshold $\tau$ is set to 0.7~\cite{ssul} for all experiments. 
For Pascal VOC 2012, the training proceeds with the batch size of 16 for 32 epochs, with the learning rate of $1\times 10^{-4}$ in incremental steps and the weight for output distillation loss $\lambda_\text{LGKD}$ is set to 25. 
On ADE20k, we train the model with the batch size of 8 for 64 epochs, adjusting the learning rate to $5\times 10^{-4}$ and with $\lambda_{\text{LGKD}}$ set to 50 in continual steps. 
Finally, the $\lambda_{\text{Ortho}}$ is adaptively set to $\frac{\vert C^{t} \vert}{\vert C^{1:t} \vert}$ for all experiments.

\subsection{Comparison with the State-of-the-arts}
\label{sec.4.2}
\subsubsection{Pascal VOC.}
Experimental results on Pascal VOC with different incremental learning settings are shown in \cref{table_pascalvoc}.
As observed, our method clearly outperforms previous state-of-the-art~(SOTA) methods in all incremental scenarios.
Particularly, we highlight the remarkable performance improvement in overlapped settings and multi-step scenarios where more objects appear as the background and the error in previous distillation methods may be propagated through multiple steps.
Specifically in \cref{table_pascalvoc}, our proposed method exceeds the previous SOTA model in a single continual step~(15-5) setting by 1.4\%p~(disjoint) and 2.7\%p~(overlapped). Furthermore, our results achieve the SOTA with a large margin, up to 1.9\%p~(disjoint) and 5.0\%p~(overlapped), in the 15-1 multi-continual benchmarking scenario. For additional experimental results in the 10-1 scenario and 5-3 scenario, we refer to the Appendix~A.1.
Such large improvements in multi-continual scenarios indicate the significance of error propagation in incremental learning caused by the background shift and how well our reliable distillation techniques, i.e., selective pseudo-labeling and adaptive feature distillation, address the background shift towards old classes.
Furthermore, relatively large improvements in overlapped scenarios compared to disjoint scenarios are due to the inclusion of future class knowledge in the background in overlapped settings.
Therefore, we believe that leveraging the background knowledge to initialize the new class token and decoupling the shared knowledge benefits the most in such settings.
\begingroup
\setlength{\tabcolsep}{2.3pt} 
\renewcommand{\arraystretch}{1.2}
\begin{table}[t]
\caption{Performance comparison on ADE20k across various scenarios in \textit{overlapped} setting. CNN and Transformer column indicates the type of the backbone network.}
\label{table_ade20k}
{
\scriptsize
\begin{tabular}{l|ccc|ccc|ccc|ccc}
\hlineB{2.5}
\multicolumn{1}{l|}{\multirow{2}{*}{Method}} & \multicolumn{3}{c|}{100-50 (2 steps)} & \multicolumn{3}{c|}{50-50 (3 steps)} & \multicolumn{3}{c|}{100-10 (6 steps)} & \multicolumn{3}{c}{100-5 (11 steps)} \\
\multicolumn{1}{c|}{} & 1-100 & 101-150 & All & 1-50 & 51-150 & All & 1-100 & 101-150 & All & 1-100 & 101-150 & All\\ \hline
\multicolumn{13}{c}{\textbf{CNN-based Methods}} \\ \hline
MiB & 40.5 & 17.2 & 32.8 & 45.5 & 21.0 & 29.3 & 38.2 & 11.1 & 29.2 & 36.0 & 5.7 & 26.0 \\
SDR & 37.4 & 24.8 & 33.2 & 40.9 & 23.8 & 29.5 & 28.9 & 7.4 & 21.7 & - & - & - \\ 
PLOP & 41.7 & 15.4 & 33.0 & 47.8 & 21.6 & 30.4 & 39.4 & 13.6 & 30.9 & 39.1 & 7.8 & 28.8 \\
REMINDER& 41.6 & 19.2 & 34.1 & 47.1 & 20.4 & 29.4 & 39.0 & 21.3 & 33.1 & - & - & -\\
RCIL & 42.3 & 18.8 & 34.5 & 48.3 & 25.0 & 32.5 & 39.3 & 17.6 & 32.0 & 38.5 & 11.5 & 29.6\\
SPPA & 42.9 & 19.9 & 35.2 & 49.8 & 23.9 & 32.5 & 41.0 & 12.5 & 31.5 & - & - & -\\
RBC & 42.9 & 21.5 & 35.8 & 49.6 & 26.3 & 34.2 & 39.0 & 21.7 & 33.3 & - & - & -\\ \cline{1-1}  \cline{1-13} 
Joint & 43.9 & 27.2 & 38.3 & 50.9 & 32.1 & 38.3 & 43.9 & 27.2 & 38.3 & 43.9 & 27.2 & 38.3 \\ \hline
\multicolumn{13}{c}{\textbf{Transformer-based Methods}} \\ \hline
MiB & 46.4 & 35.0 & 42.6 & 52.2 & 35.6 & 41.1 & 43.0 & 30.8 & 38.9 & 40.2 & 26.6 & 35.7 \\
INC & 49.4 & 35.6 & 44.8 & 56.2 & 37.8 & 43.9 & 48.5 & 34.6 & 43.9 & 46.9 & 31.3 & 41.7 \\
\rowcolor{gray!30}
\textbf{Ours} & \textbf{50.0} & \textbf{37.1} & \textbf{45.7} & \textbf{57.0} & \textbf{39.7} & \textbf{45.4} & \textbf{49.0} & \textbf{35.4} & \textbf{44.5} & \textbf{48.0} & \textbf{32.3} & \textbf{42.8} \\ \cline{1-1}  \cline{1-13}
Joint & 50.1 & 41.4 & 47.2 & 57.6 & 42.0 & 47.2 & 50.1 & 41.4 & 47.2 & 50.1 & 41.4 & 47.2 \\ \hlineB{2.5}
\end{tabular}
}
\end{table}
\endgroup
\subsubsection{ADE20k.} 
Our experiments in the overlapped setting on ADE20k dataset are reported in \cref{table_ade20k}. 
As demonstrated, we achieve an average improvement of 1.0\%p over the state-of-the-art methods in the 100-50, 50-50, 100-10, and 100-5 scenarios.
In particular, as the scenarios become more challenging, \textit{i.e.}, more time steps, our proposed method showcases its benefit in retaining its knowledge~(stability) beyond the previous SOTA.
Specifically, the margins between the performances for the classes in the initial step are 0.6\%p in 100-50~(2 steps), 0.8\%p in 50-50~(3 steps) and 1.1\%p in 100-5~(11 steps) while also exceeding the performances for measuring the new class learning.
Overall, these results verify the effectiveness of ours in both the stability and the plasticity perspective which demonstrates the strength in CISS.

\subsection{Ablation Studies}
\label{sec.4.3}
\subsubsection{Component Analysis.}
We investigate the effectiveness of each component in the overlapped setting of Pascal VOC 15-1 and ADE20k 100-50, as observed in \cref{table_ablation}.
Compared to the baseline~(a), rows (b) to (c) show the benefits of SPL and AFD discussed in \cref{sec.3.3} and \cref{sec.3.4}.
Specifically, the performances for the classes in the initial step~(1-15) are enhanced by 2.4\%p and 1.4\%p in the 15-1 scenario.
We owe this improvement to their capability of preventing the old class pixels from being labeled as the background.
Surprisingly, we also find that SPL and AFD also benefit in classes that appear in later steps, SPL particularly showing a 13.2\%p increase.
This is because selective and adaptive strategies can also suppress the learning of future class objects with the background label if they appear in earlier steps.
On the other hand, our objective $\mathcal{L}_\text{Sep}$, explained in \cref{sec.3.5}, enhances the model's plasticity~(result for new classes in row (d) is 4.5\%p higher than the baseline~(a) in Pascal VOC 15-1 setting) by directly separating the embeddings of the new classes from the background.
Consequently, the results in (h) with all components combined demonstrate not only the effectiveness of each component but also the compatibility to one another.
\begingroup
\setlength{\tabcolsep}{6pt} 
\renewcommand{\arraystretch}{1.2}
\begin{table}[t]
\caption{Ablation study for each component on Pascal VOC 15-1 and ADE20k 100-50 in \textit{overlapped} setting. 
     SPL, AFD, $\mathcal{L}_{\text{Sep}}$ denote selective pseudo-labeling, adaptive feature distillation, and separating objective, respectively.
    }
    \label{table_ablation}
	\centering
	{\scriptsize
    \begin{tabular}{c|ccc|ccc|ccc}
    \hlineB{2.5}
    & \multirow{3}{*}{SPL} & \multirow{3}{*}{AFD} & \multirow{3}{*}{$\mathcal{L}_{\text{Sep}}$} & \multicolumn{3}{c|}{Pascal VOC} & \multicolumn{3}{c}{ADE20k} \\ \cline{5-10}
    &  &  &  & \multicolumn{3}{c|}{15-1 (6 steps)} & \multicolumn{3}{c}{100-50 (2 steps)} \\ 
    &  &  &  & 1-15 & 16-20 & All & 1-100 & 101-150 & All \\  \hline
    (a) & - & - &  - & 81.1 & 49.8 & 74.2 & 47.3 & 35.0 & 43.2 \\ 
    (b) & \checkmark & - & - & 83.5 & 63.0 & 79.1 & 50.0 & 35.5 & 45.2\\ 
    (c) & - & \checkmark & - & 82.5 & 57.3 & 77.0 & 47.4 & 36.4 & 43.7\\ 
    (d) & - & - & \checkmark & 80.8 & 69.6 & 78.7 & 47.1 & 37.3 & 43.8\\
    (e) & \checkmark & \checkmark & - & \textbf{83.8} & 65.3 & 79.9 & 50.0 & 35.7 & 45.2\\
    (f) & \checkmark & - & \checkmark & 82.2 & 70.2 & 79.9 & 49.7 & 36.9 & 45.5\\ 
    (g) & - & \checkmark & \checkmark & 81.2 & 70.8 & 79.3 & 47.4 & \textbf{37.6} & 44.1\\ \rowcolor{gray!30}
    (h) & \checkmark & \checkmark & \checkmark & 82.6 & \textbf{72.2} & \textbf{80.6} & \textbf{50.0} & 37.1 & \textbf{45.7}\\
    \hlineB{2.5}
    \end{tabular}
    }
\end{table}
\endgroup

\begingroup
\setlength{\tabcolsep}{6pt} 
\renewcommand{\arraystretch}{1.5}
\begin{table}[t]
\caption{Ablation study for selective pseudo-labeling filter effectiveness on Pascal VOC 15-1 in \textit{overlapped} setting. 
    }
    \label{SPL_ablation}
	\centering
	{\scriptsize
    \begin{tabular}{c|cccccc}
    \hlineB{2.5}
    & \textls[-60]{Step 1} & \textls[-60]{Step 2} & \textls[-60]{Step 3} & \textls[-60]{Step 4} & \textls[-60]{Step 5} & Avg. \\ \cline{1-7}
    Ignored Label Ratio & 38.21 & 31.29 & 34.59 & 34.32 & 47.52 & 37.18 \\ 
    \hlineB{2.5}
    \end{tabular}
    }
\end{table}
\endgroup
\subsubsection{Effectiveness of Selective Strategy within Pseudo-labeling.}
\setlength{\columnsep}{6pt}
\setlength{\intextsep}{3pt}
\renewcommand{\arraystretch}{1.2}
\begin{wraptable}{r}{0.5\textwidth} 
  \caption{Ablation study evaluating the impact of the Orthogonal Objective in the ADE20k 100-50 \textit{overlapped} setting.}
  \label{ablation_similarity}
  \centering
  {\scriptsize
    \begin{tabular}{c|cc|ccc}
     \hlineB{2.5}
     & \textls[-50]{\multirow{2}{*}{$\mathcal{L}_{\text{LGKD}}$}} & \textls[-50]{\multirow{2}{*}{$\mathcal{L}_{\text{Ortho}}$}} &\multicolumn{3}{c}{Similarity} \\ 
     & & & \textls[-70]{($C^t$, $C^t$)} & \textls[-70]{($C^t$, $c_0$)} & \textls[-70]{($C^t$, $C^{1:t-1}$)} \\ \cline{1-6}
     (a) & - & - & 91.24 & 91.46 & 3.03 \\
     (b) & \checkmark & - & 90.15 & 90.03 & 2.93 \\ \rowcolor{gray!30}
     (c) & \checkmark & \checkmark & 52.68 & 53.86 & 1.58 \\ 
     \hlineB{2.5}
     \end{tabular}
  }
\end{wraptable}
We inspect how much the incorrectly labeled pixels are detected with our selective strategy.
In \cref{SPL_ablation}, we present the percentage of the old class pixels identified by selective strategy among mispredicted pixels by na\"ive pseudo-labeling for each step.
Briefly, our selective method filters out about 37\% mispredicted old class pixels as the background on average.
In other words, we claim that being selective within pseudo-labeling can alleviate the background shift towards old classes up to 37\% which demonstrates the strengths of our method.
\begin{figure}[t]
    \centering
    \includegraphics[width=1.0\textwidth]{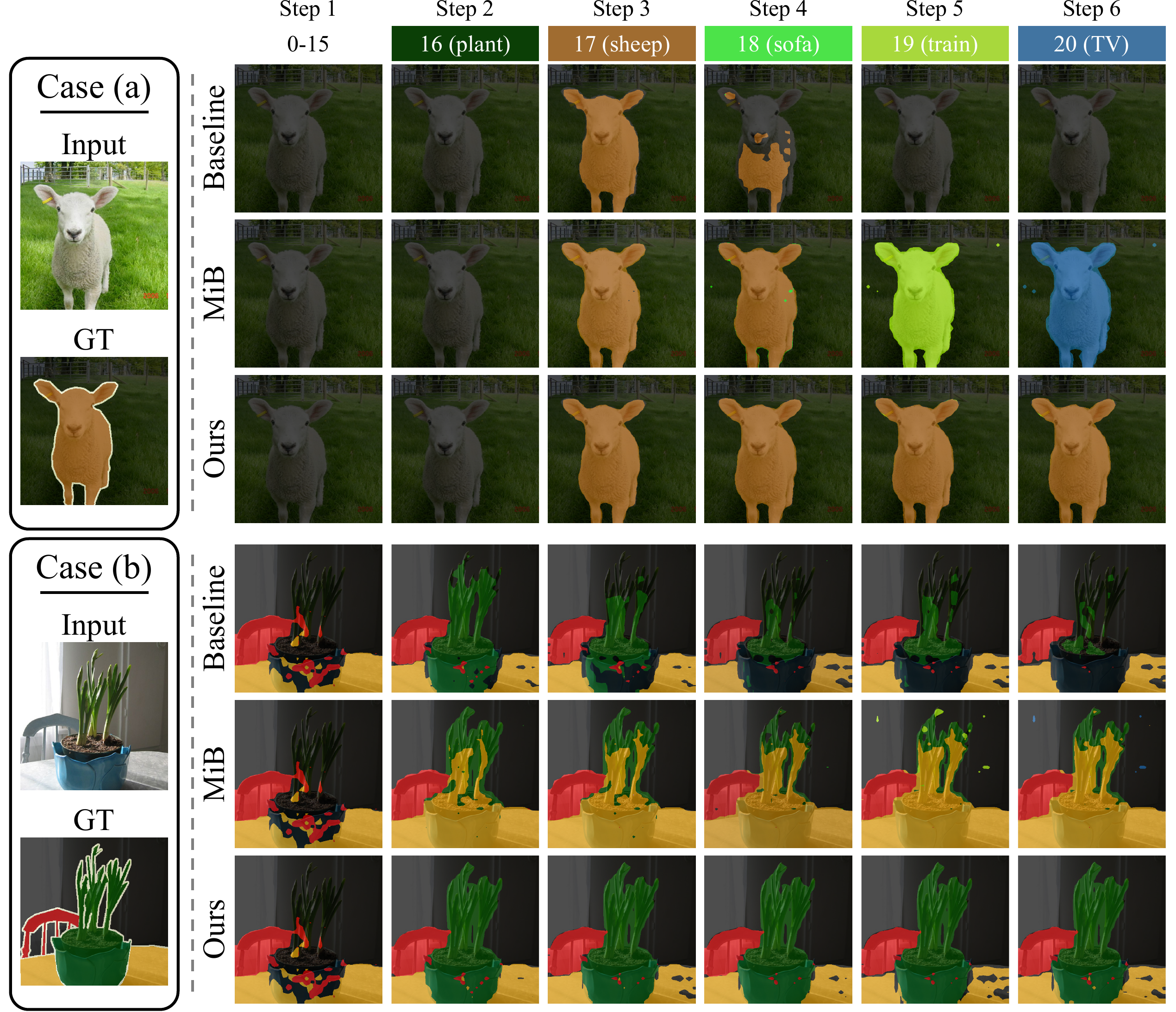}
    \caption{Comparison of qualitative results on the 15-1 protocol of the Pascal VOC between the baseline, MiB, and ours.}
    \label{fig:Qualitative}
\end{figure}
\subsubsection{Class Token Similarity.}
To further validate the effect of separating objective, we present the class token similarities, that more directly represent the degree of separation between classes, in \cref{ablation_similarity}.
As shown in row~(a), when only the background transfer is used, the similarities for new class pairs~($C^t$, $C^t$) and new-to-background~($C^t$, $c_0$) are exceptionally high at 91.24 and 91.46, respectively. 
Even if the label-guided knowledge distillation $L_{\text{LGKD}}$~\cite{LGKD} is adopted, it shows a marginal separation as like 90.15 and 90.03, individually.
On the other hand, when our separating objective $L_{\text{Ortho}}$ is utilized together, the similarity scores extremely decrease to 52.68 and 53.86, respectively.
Moreover, the similarity between new-to-old~($C^t$, $C^{1:t-1}$) decreased to 1.58.
This proves that the orthogonal objective possessing a direct separating purpose is effective in isolating the new classes from the background class and mitigates the background shift towards new classes.

\begin{figure}[t]
    \centering
    \includegraphics[width=1.0 \textwidth]{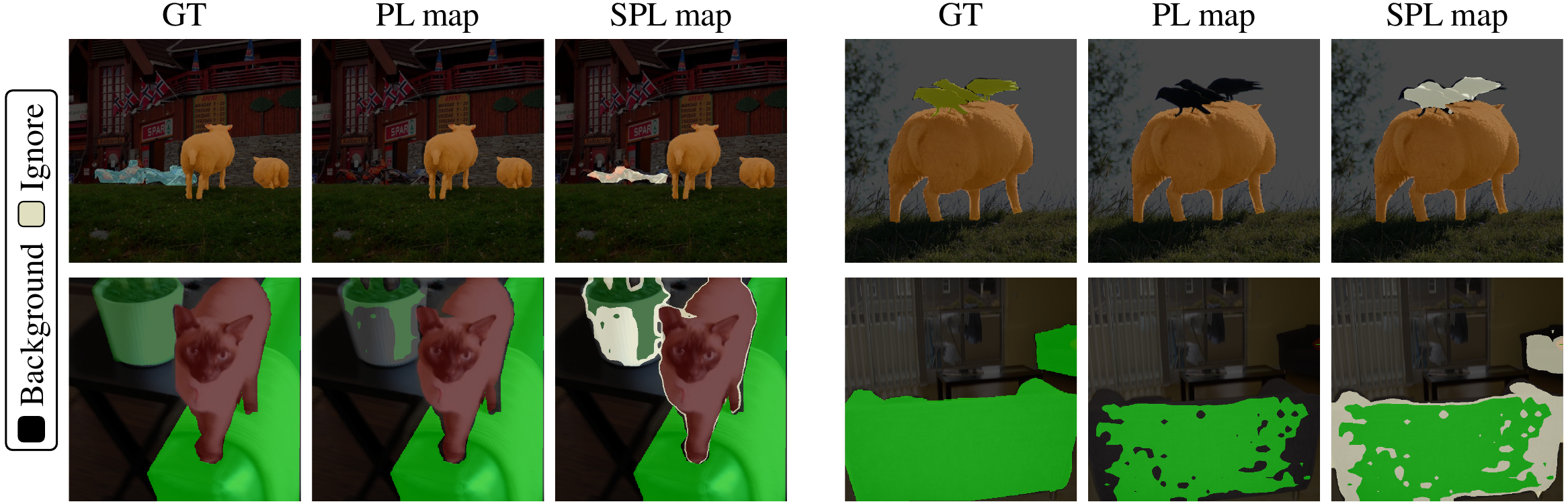}
    \caption{Comparisons between the Pseudo-Labeling map~(PL map) and the Selective Pseudo-Labeling map~(SPL map). Both of them are yielded from our method.}
    \label{fig:object_idenfier}
\end{figure}
\subsection{Qualitative Analysis}
We visualize the qualitative results in the Pascal VOC 15-1 \textit{overlapped} setting in \cref{fig:Qualitative}. 
Across examples, our proposed methods show its benefits in terms of robustness to forgetting and plasticity to learning new classes compared to both the baseline as mentioned in \cref{sec.4.3} and MiB.
For example, whereas the baseline and MiB struggle to retain the old knowledge~(a-step~5, and b-step~6), our method successively captures the old classes.
Furthermore, (b-step~2) display difficulty in learning new classes in the baseline and MiB, respectively, while our methods are shown to learn the new class `plant' precisely.
These results show that our method possesses robustness to background shifts towards old and new classes.
The additional qualitative results can be found in the Appendix.

To show the role of the object identifier $O$, we visualize the na\"ive pseudo-labeling and selective pseudo-labeling maps in~\cref{fig:object_idenfier}.
The na\"ive pseudo-labeling map shows its vulnerability to the imprecision of the old model, labeling the ambiguous objects as the background.
In contrast, our selective pseudo-labeling alleviates the background shift by detecting and filtering out those objects.


\section{Conclusion}
In this paper, we propose a background-class separation framework to address the background shift issue towards the old and new classes in CISS. To alleviate the prediction bias towards the background in old classes, we first introduced selective pseudo labeling and adaptive feature distillation that are to scale down the degree of distillation in parts where the false knowledge might be distilled.
Subsequently, to facilitate the decoupling of background and new classes while leveraging the benefits of the background weight transfer strategy, we encourage further separation between the background and new classes. This is achieved through a novel orthogonal objective, complemented by label-guided output distillation.
With all our components combined, our proposed method achieved state-of-the-art performances in terms of both robustness to forgetting and the plasticity of new class learning.

\noindent\textbf{Acknowledgements.} This work was supported in part by MSIT\&KNPA/KIPoT (Police Lab 2.0, No. 210121M06), MSIT/IITP (No. 2022-0-00680, 2019-0-00421, 2020-0-01821, RS-2024-00437102), and SEMES-SKKU collaboration funded by SEMES. Gilhan Park acknowledges support from the Hyundai Motor Chung Mong-Koo Foundation.
\renewcommand{\thesection}{A}   
\renewcommand{\thetable}{A\arabic{table}}   
\renewcommand{\thefigure}{A\arabic{figure}}
\setcounter{section}{0}
\setcounter{table}{0}
\setcounter{figure}{0}
\section{Appendix}
\setlength{\columnsep}{5pt} 
\setlength{\intextsep}{5pt} 
\setlength{\tabcolsep}{5pt} 
\renewcommand{\arraystretch}{1.0} 
\begin{table}
    \caption{Results on Pascal VOC 10-1~(11 steps) and 5-3~(6 steps) \textit{overlapped} setting. ${\dagger}$ indicates the reproduced results from SSUL\cite{ssul}.}
    \label{table_pascalvoc5-3}
    \centering
    	{
        \footnotesize
        \begin{tabular}{l|ccc|ccc}
        \hlineB{2.5}
        \multirow{2}{*}{Method}& \multicolumn{3}{c|}{10-1~(11 steps)} & \multicolumn{3}{c}{5-3~(6 steps)} \\ 
        & 1-10 & 11-20 & All & 1-5 & 6-20 & All \\  \hline
        \multicolumn{7}{c}{\textbf{CNN-based Methods}} \\ \hline
        ILT$^{\dagger}$~\cite{ILT} & 7.1 & 3.6 & 5.5 & 22.5 & 31.6 &  29.0 \\ 
        MiB$^{\dagger}$~\cite{MiB} & 20.0 & 20.1 & 20.1 & 57.1 & 42.6  & 46.7 \\
        SDR~\cite{SDR} & 32.4 & 17.1  & 25.1 & - & - & - \\
        PLOP$^{\dagger}$~\cite{PLOP} & 44.0 & 15.5 & 30.5 & 17.5 & 19.2 & 18.7 \\ 
        RECALL~\cite{RECALL} & 59.5 & 46.7 & 54.8 & - & - & - \\
        RCIL~\cite{RC} & 55.4 & 15.1 &  34.3 & - & - & - \\ \hline
        Joint & 79.8 & 72.6 & 78.2 & 76.9 & 77.6 & 77.4 \\ \hline
        \multicolumn{7}{c}{\textbf{Transformer-based Methods}} \\ \hline
        MiB~\cite{MiB} & - & - & - & 33.4 & 43.2 & 42.9 \\
        INC~\cite{Incrementer} & 77.62 & 60.33 & 70.16 & - & - & - \\
        \rowcolor{gray!30}
        \textbf{Ours} & \textbf{80.95} & \textbf{71.98} & \textbf{77.19} & \textbf{77.5} & \textbf{77.5} & \textbf{78.1} \\ \hline
        Joint & 83.33 & 84.31 & 83.81 & 79.6 & 85.2 & 83.8 \\
        \hlineB{2.5}
        \end{tabular}
        }
\end{table}
\subsection{Additional quantitative results}
\subsubsection{Scenario Study with Small Number of Inital Classes.}
The experimental results for scenarios with a small initial class set are reported in \cref{table_pascalvoc5-3}.
Experiments are conducted with Pascal VOC dataset.
The 10-1 steps and 5-3 steps configuration present a more challenge due to its increased susceptibility to forgetting.
This challenge arises from the initial step having a limited class set, in contrast to the subsequent steps which encompass a broader class set. 
In such a difficult scenario, our method shows consistent results, achieving a significant performance improvement.
Specifically, in the scenario of steps 5-3,our method exceeds MiB~\cite{MiB} by 44.04\%p in the initial step~(1-5) and 34.24\%p in the continual steps~(6-20). Additionally, in the 10-1 scenario, the initial step~(1-10) was 3.33\%p higher and the continuous steps~(11-20) were 11.65\%p higher than the previous state-of-the-art (SOTA) model, INC~\cite{Incrementer}.
This demonstrates the effectiveness of our distillation techniques, i.e., selective pseudo-labeling and adaptive feature distillation, in significantly addressing the background shift towards old classes. Furthermore, the improved performance in subsequent steps with a more extensive class set demonstrates the advantages of leveraging background knowledge to initialize new class tokens and the decoupling of shared knowledge, particularly in these scenarios. This approach effectively addresses the background shift towards new classes.
\subsubsection{Ablation Study in the Disjoint setting.}
As discussed in Sec.~4.1, the disjoint setting is relatively simpler challenges compared to the overlapped setting, as future class objects do not appear in the images of the current step.
In this subsection, we further investigate each component within the disjoint setting as shown in \cref{table_ablation_disjoint}.
In short, we find that an ablation study within a disjoint setting consistently validates the effectiveness of all components.
\begingroup
\setlength{\tabcolsep}{10pt} 
\renewcommand{\arraystretch}{1.}
\begin{table}[t]
\caption{Ablation study for each component on Pascal VOC 15-1 in \textit{disjoint} setting. 
     SPL, AFD, $\mathcal{L}_{\text{Sep}}$ denote Selective Pseudo-Labeling, Adaptive Feature Distillation, and Separating objective, respectively.
    }
    \label{table_ablation_disjoint}
	\centering
	{
    \begin{tabular}{c|ccc|ccc}
    \hlineB{2.5}
    & \multirow{3}{*}{SPL} & \multirow{3}{*}{AFD} & \multirow{3}{*}{$\mathcal{L}_{\text{Sep}}$} & \multicolumn{3}{c}{Pascal VOC} \\ \cline{5-7}
    &  &  &  & \multicolumn{3}{c}{15-1 (6 steps)} \\ 
    &  &  &  & 1-15 & 16-20 & All \\  \hline
    (a) & - & - &  - & 77.5 & 54.4 & 72.7 \\ 
    (b) & \checkmark & - & - & 81.0 & 58.7 & 76.3 \\ 
    (c) & - & \checkmark & - & 78.3 & 54.9 & 73.4 \\ 
    (d) & - & - & \checkmark & 79.1 & 61.6 & 75.6 \\
    (e) & \checkmark & \checkmark & - & 81.4 & 59.7 & 76.9 \\
    (f) & \checkmark & - & \checkmark & 80.9 & 63.7 & 77.4 \\ 
    (g) & - & \checkmark & \checkmark & 79.1 & 62.1 & 75.7 \\
    (h) & \checkmark & \checkmark & \checkmark & \textbf{81.5} & \textbf{64.7} & \textbf{78.1}\\
    \hlineB{2.5}
    \end{tabular}
    }
\end{table}
\endgroup

\begingroup
\setlength{\tabcolsep}{4pt}
\renewcommand{\arraystretch}{1.}
\begin{table}
\caption{Results on Pascal VOC 15-5~(2 steps) and 15-1~(6 steps) in the \textit{overlapped} setting, along with ADE20k 100-10~(6 steps) in the \textit{overlapped} setting, all utilizing a CNN backbone. ${\dagger}$ indicates the reproduced results from \cite{PLOP, RBC, Incrementer}}
\label{table_CNN}
\centering
    {
    \begin{tabular}{l|ccc|ccc|ccc}
    \hlineB{2.5}
    \multicolumn{1}{c|}{\multirow{3}{*}{}} & \multicolumn{6}{c|}{Pascal VOC} & \multicolumn{3}{c}{ADE20k}\\ \cline{2-10}
    & \multicolumn{3}{c|}{15-5~(2 steps)} & \multicolumn{3}{c|}{15-1~(6 steps)} & \multicolumn{3}{c}{100-10~(6 steps)} \\ \cline{2-10}
    & \multicolumn{1}{c}{1-15} & \multicolumn{1}{c}{16-20} & \multicolumn{1}{c|}{All} & \multicolumn{1}{c}{1-15} & \multicolumn{1}{c}{16-20} & \multicolumn{1}{c|}{All} & \multicolumn{1}{c}{1-100} & \multicolumn{1}{c}{101-150} & \multicolumn{1}{c}{All} \\ \hline
    \multicolumn{1}{c}{} & \multicolumn{9}{c}{\textbf{CNN-based Methods}} \\ \hline
    MiB$^{\dagger}$~\cite{MiB} & 76.4 & 50.0 & 70.1 & 34.2 & 13.5 & 29.3 & 38.2 & 11.1 & 29.2 \\
    SDR$^{\dagger}$~\cite{SDR} & 75.4 & 52.6 & 69.9 & 44.7 & 21.8 & 39.2 & 28.9 & 7.4 & 21.7 \\
    PLOP$^{\dagger}$~\cite{PLOP} & 75.7 & 51.7 & 70.1 & 65.1 & 21.1 & 54.6 & 39.4 & 13.6 & 30.9 \\
    REMIND~\cite{REMIND} & 76.1 & 50.7 & 70.1 & 68.3 & 27.2 & 58.5 & 39.0 & 21.3 & 33.1 \\
    RCIL~\cite{RC} & 78.8 & 52.2 & 72.4 & 70.6 & 23.7 & 59.4 & 39.3 & 17.6 & 
 32.0\\
    SPPA~\cite{SPPA} & 78.1 & 52.9 & 72.1 & 66.2 & 23.3 & 56.0 & 41.0 & 12.5 & 31.5 \\
    RBC$^{\dagger}$~\cite{RBC} & 76.6 & 52.8 & 70.9 & 69.5 & 38.4 & 62.1 & 39.0 & 21.7 & 33.3 \\
    \rowcolor{gray!30}
    \textbf{Ours} & \textbf{78.8} & \textbf{54.3} & \textbf{73.3} & \textbf{76.1} & \textbf{39.1} & \textbf{67.8} & \textbf{42.8} & \textbf{22.6} & \textbf{36.1} \\ \cline{1-1}  \cline{1-10}
    Joint & 79.1 & 72.6 & 77.4 & 79.1 & 72.6 & 77.4 & 43.9 & 27.2 & 38.3 \\
    \hlineB{2.5}
\end{tabular}
}
\end{table}
\endgroup

\subsubsection{Results with CNN Backbone.}
We evaluate the effectiveness of our method using a CNN architecture. Specifically, we utilize a DeepLabv3~\cite{deeplabv3} segmentation network with a ResNet-101~\cite{resnet} backbone, which is pre-trained on ImageNet~\cite{imagenet}. 
For other configurations, we follow PLOP~\cite{PLOP}. 
We evaluate our method on two datasets: Pascal VOC~\cite{pascalvoc} and ADE20k~\cite{ade20k}. 
In the Pascal VOC overlapped setting, experiments are conducted in the 15-5~(2 steps) and 15-1~(6 steps) scenarios, and similarly, in the ADE20k dataset overlapped setting, we perform experiments in the 100-10~(6 steps). As shown in \cref{table_CNN}, our method clearly outperforms all previous methods, demonstrating the strengths also with CNN backbones.

\subsubsection{Hyperparameter Ablation.}
A threshold $\tau$ is set on the output of the previous model to generate pseudo-labels. 
To compare between different strategies of selecting hyperparameter $\tau$, we conduct ablation studies with different pseudo-labeling methods, such as employing a sigmoid function~\cite{ssul} and using entropy function~\cite{PLOP}. 
We reported the results of threshold selection in \cref{weights_and_effectiveness_study}~(right). 
As observed, we find that setting a threshold $tau$ to 0.7 is suitable for continual scenarios, ensuring the reliability of pseudo-labels thereby reducing the number of noisy labels.
Additionally, we also point out the robustness of our method to different constant $\tau$ values.
$\lambda_{\text{LGKD}}$~(ranging from 5 to 50) and $\lambda_{\text{ortho}}$~(from 0.5 to 0.01) are also examined in the left and the middle columns.
These results further demonstrate the robustness of the model to hyperparameter variations, showing the consistent performances across different hyperparameter values.

\begingroup
\setlength{\tabcolsep}{3pt} 
\renewcommand{\arraystretch}{1.}
\begin{table}[t]
\caption{Ablation on Pascal VOC 15-1 \textit{overlapped} setting to evaluate the effect of hyperparameters.}
\label{weights_and_effectiveness_study}
	\centering
	{
    \begin{tabular}{c|ccc|c|ccc|c|ccc}
    \hlineB{2.5}
     $\lambda_{\text{LGKD}}$ & 1-15 & 16-20 & All & $\lambda_{\text{Ortho}}$ & 1-15 & 16-20 & All & Thr.~$\tau$ & 1-15 & 16-20 & All \\  \hline
    5 & 82.0 & 69.1 & 79.5 & 0.5 & 82.4 & 70.2 & 80.1 & 0.6 & 82.6 & 71.9 & 80.5 \\ 
    10 & 82.3 & 70.8 & 80.1 & 0.1 & 82.3 & 71.5 & 80.2 & \textbf{0.7} & \textbf{82.6} & \textbf{72.2} & \textbf{80.6} \\ 
    20 & 82.6 & 71.4 & 80.4 & 0.05 & 82.5 & 72.2 & 80.5 & 0.8 & 82.4 & 71.3 & 80.2 \\ 
    \textbf{25} & \textbf{82.6} & \textbf{72.2} & \textbf{80.6} & 0.01 & 82.5 & 72.1 & 80.5 & PLOP & 79.6 & 68.2 & 77.4 \\
    50 & 82.3 & 70.14 & 79.9 & \tiny\textbf{${\vert C^{t} \vert}/{\vert C^{1:t} \vert}$} &\textbf{82.6} & \textbf{72.2} & \textbf{80.6} & SSUL & 81.5 & 70.0 & 79.3\\ 
    \hlineB{2.5}
    \end{tabular}
    }
\end{table}
\endgroup

\subsection{Additional qualitative results}
\subsubsection{Additional qualitative results for SPL.}
\begin{figure}
    \centering
    \includegraphics[width=1. \textwidth]{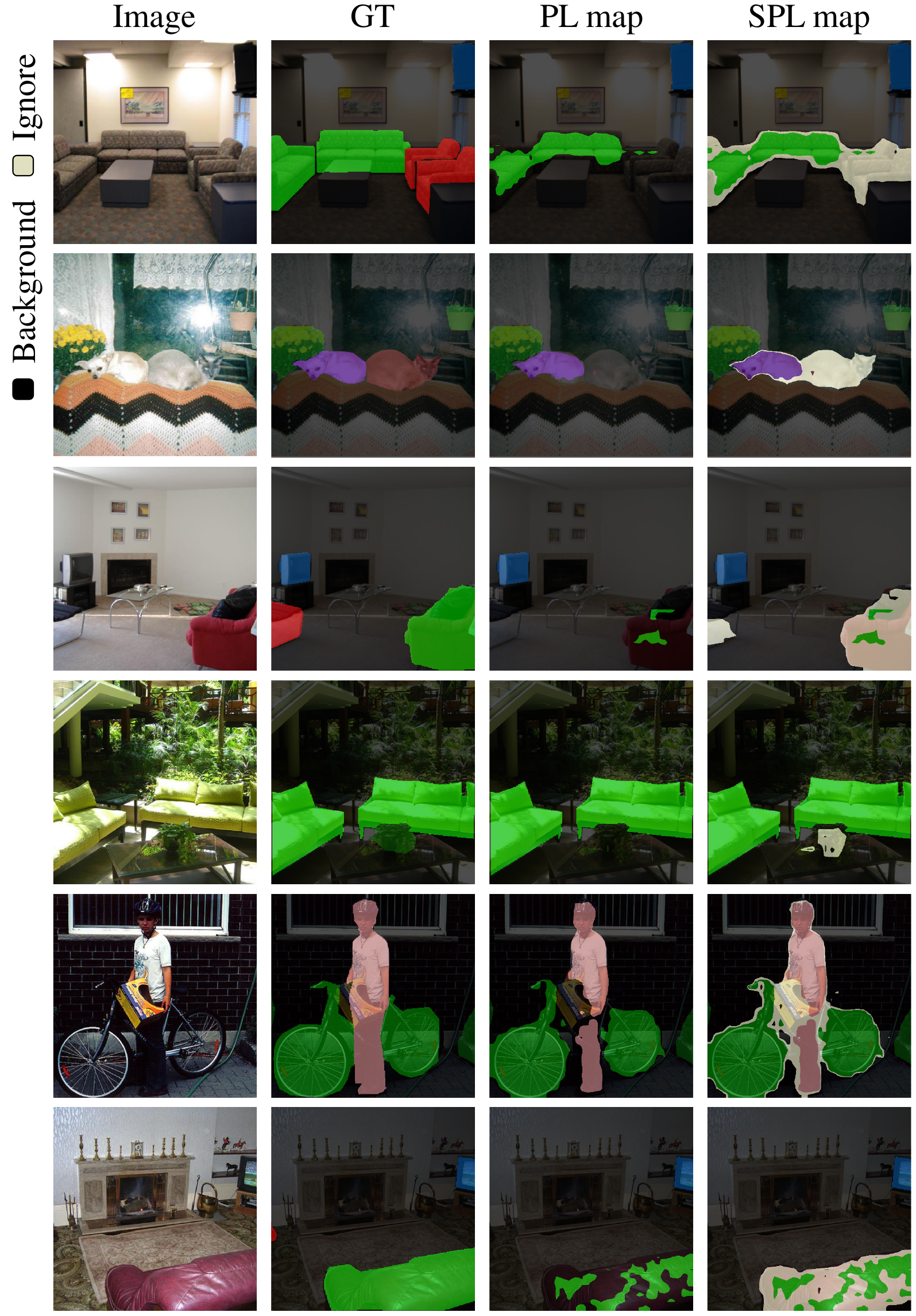}
    \caption{Visualization of Pseudo-Labeling~(PL) map and Selective Pseudo-Labeling~(SPL) map.
    By comparing these two, we highlight the primary objective of selective strategies in pseudo-labeling: to effectively exclude pixels that might be mislabeled from the training process.}
    \label{fig:object_idenfier_supple}
\end{figure}

\cref{fig:object_idenfier_supple} shows additional visualization results of the object identifier $O$ which is introduced in Sec.~3.3. 
These results illustrate the weakness of conventional pseudo-labeling methods; ambiguous pixels, \textit{e.g.}, sofa, chair, plant, and cat, are labeled as background due to the inaccuracy of the old model. However, our selective pseudo labeling using object identifier $O$ is shown to be capable of detecting ambiguous pixels.

\subsubsection{Additional qualitative results on Pascal VOC.} 
\begin{figure}
    \centering
    \includegraphics[width=1. \textwidth]{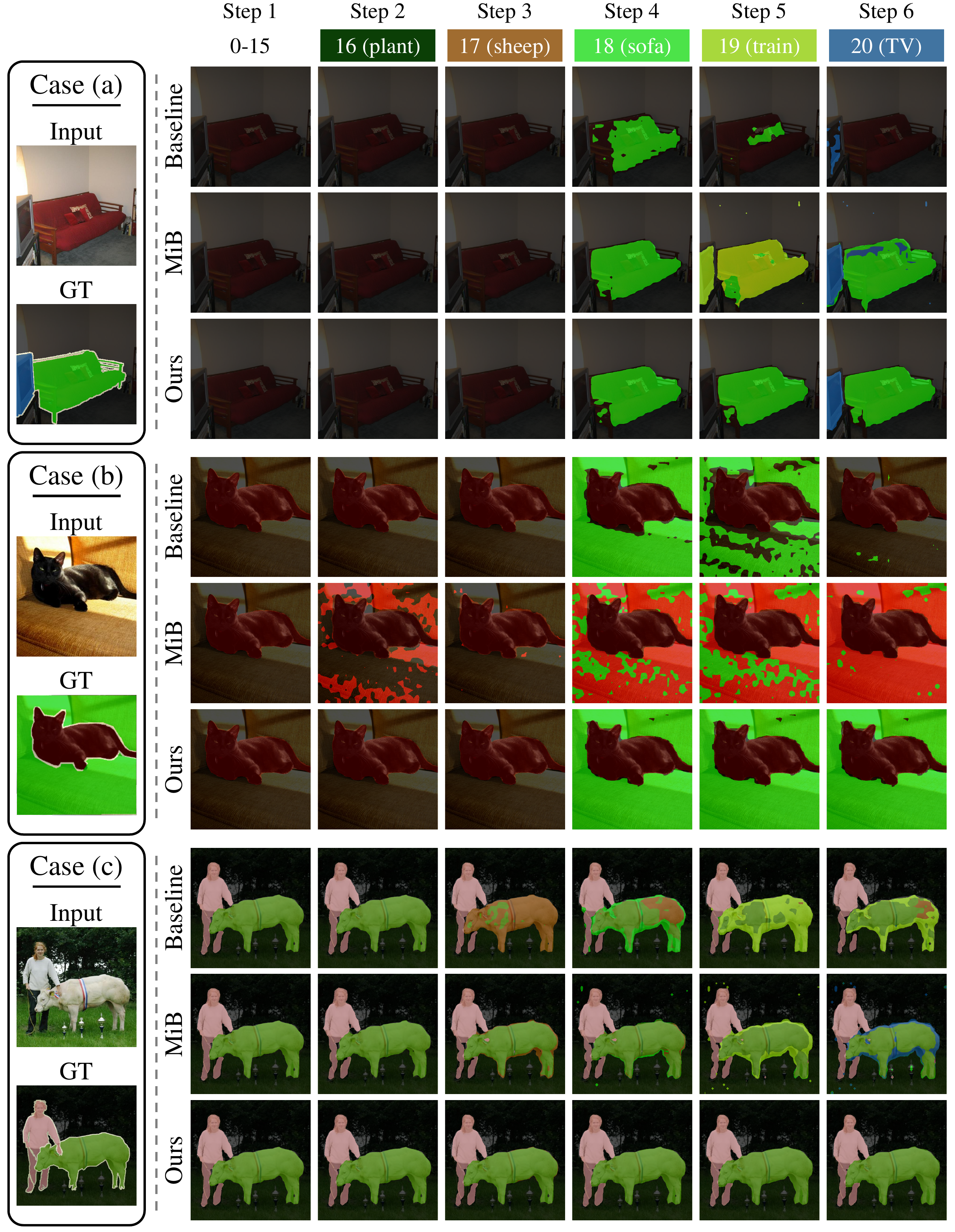}
    \caption{Comparison of qualitative results on the 15-1 protocol of the Pascal VOC between the baseline, MiB, and ours.}
    \label{fig:qualitative_results_supple_v1}
\end{figure}

We present additional qualitative results in the Pascal VOC 15-1 \textit{overlapped} setting in \cref{fig:qualitative_results_supple_v1}. Our proposed methods demonstrate their advantages in terms of robustness against forgetting and platsticity in learning new classes, outperforming both our baseline~(detailed in Sec.~4.3 and MiB~\cite{MiB}. 
For instance, while the baseline and MiB encounter difficulties in retaining old knowledge, often resulting in a background shift or overfitting to new classes~(a-step 5, b-step 6, and c-step 3), our method consistently recognizes old classes. 
Furthermore, in contrast to our method, which correctly learns the new class 'sofa', we find that MiB~\cite{MiB} (especially b-step 4) has difficulty learning the new class. These results confirm the effectiveness of our method in preventing background shifts towards both old and new classes.

\subsubsection{Qualitative results on ADE20k.} 
\begin{figure}
    \centering
    \includegraphics[width=1. \textwidth]{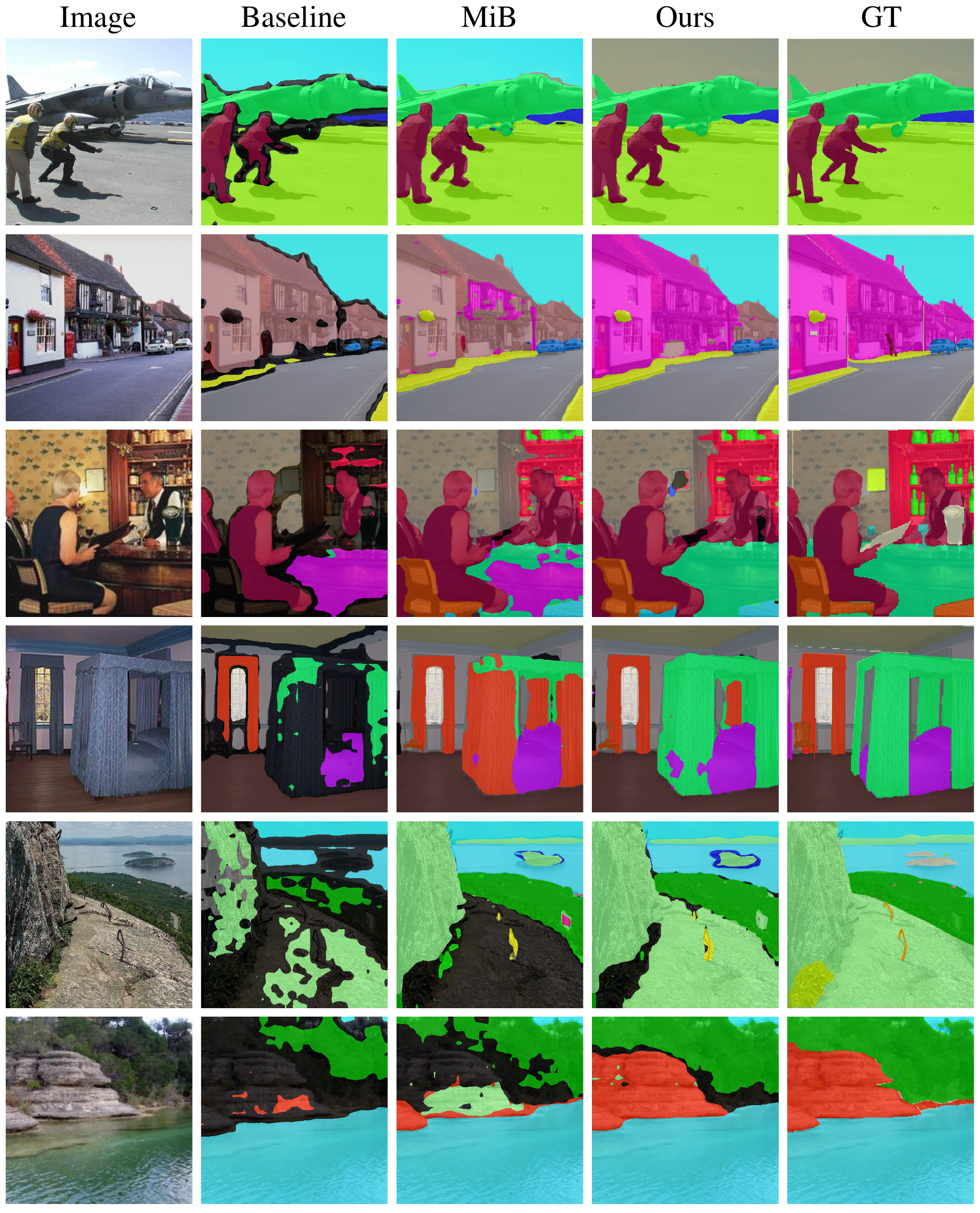}
    \caption{Qualitative results on the 100-10 protocol of the ADE20k.}
    \label{fig:ade20k_supple_1}
\end{figure}

\begin{figure}
    \centering
    \includegraphics[width=1. \textwidth]{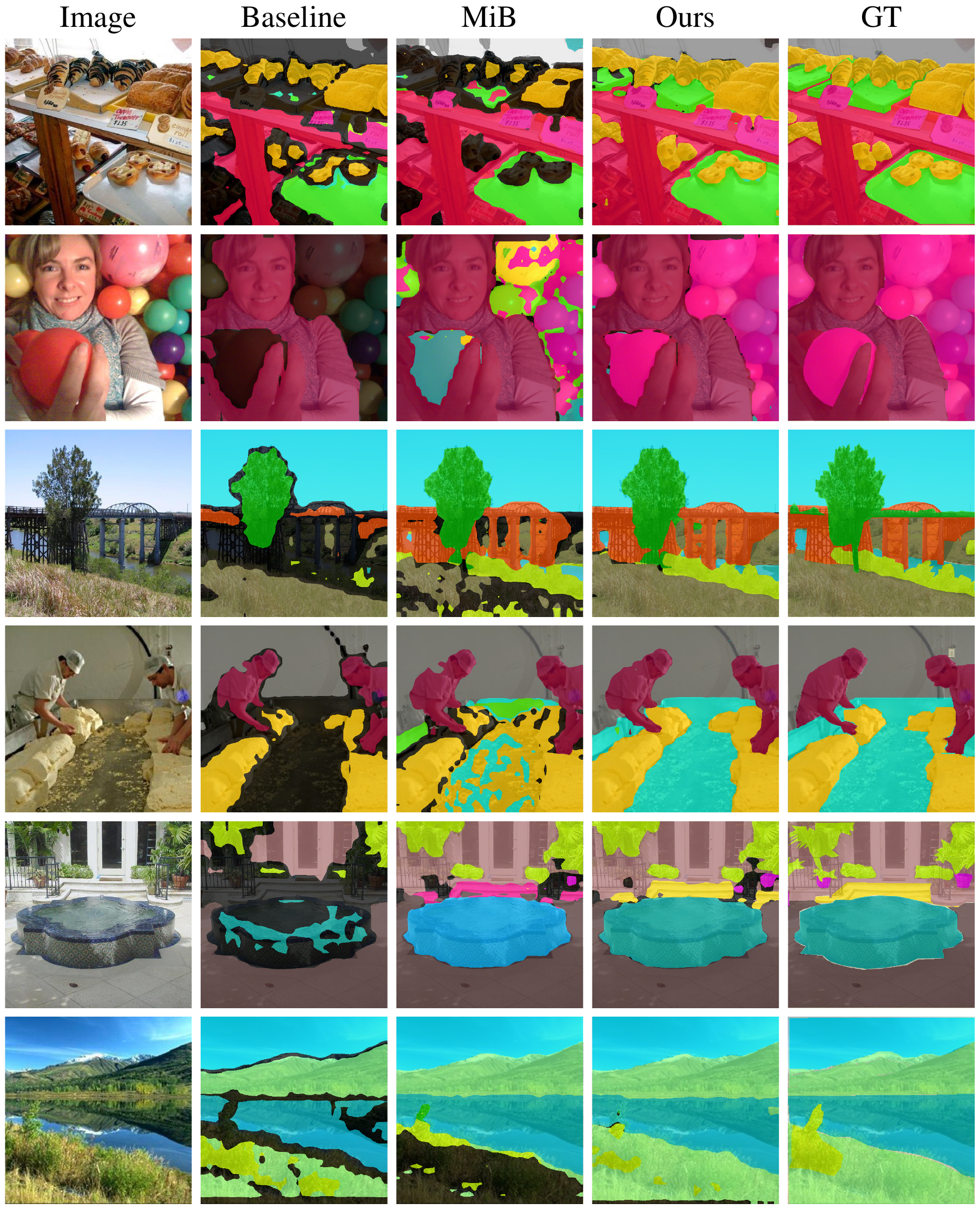}
    \caption{Qualitative results on the 100-10 protocol of the ADE20k.}
    \label{fig:ade20k_supple_2}
\end{figure}

\cref{fig:ade20k_supple_1} and \cref{fig:ade20k_supple_2} show qualitative results of our method, baseline, and MiB~\cite{MiB} on the 100-10 setting on ADE20k~\cite{ade20k}. 
The models at the final training step are used to produce visualized results. 
The baseline uses conventional pseudo-labeling, vanilla feature distillation, and output distillation. 
These techniques are detailed in Sec.~4.3.
As can be observed, both the baseline and MiB are vulnerable to background shifts, often misclassifying objects as the background class~(black) or as other object classes that are visually similar~(\textit{e.g.}, house as building). 

On the other hand, we find that our method is capable of preserving old class knowledge, as our distillation methods alleviate error propagation throughout the continual steps, effectively addressing the background shift towards old classes.

\subsection{Limitation}
Our method, particularly the selective pseudo-labeling strategy, effectively classifies ambiguous pixels for old classes, yet further development is needed for future classes. 
We believe that attempting to address limitations through analysis of our method will help direct future research on the CISS problem.

%
%
\newpage
\bibliographystyle{splncs04}
\bibliography{main}

\begin{thebibliography}{10}
\providecommand{\url}[1]{\texttt{#1}}
\providecommand{\urlprefix}{URL }
\providecommand{\doi}[1]{https://doi.org/#1}

\bibitem{MiB}
Cermelli, F., Mancini, M., Bulo, S.R., Ricci, E., Caputo, B.: Modeling the background for incremental learning in semantic segmentation. In: Proceedings of the IEEE/CVF Conference on Computer Vision and Pattern Recognition. pp. 9233--9242 (2020)

\bibitem{ssul}
Cha, S., Yoo, Y., Moon, T., et~al.: Ssul: Semantic segmentation with unknown label for exemplar-based class-incremental learning. Advances in neural information processing systems  \textbf{34},  10919--10930 (2021)

\bibitem{riemann}
Chaudhry, A., Dokania, P.K., Ajanthan, T., Torr, P.H.: Riemannian walk for incremental learning: Understanding forgetting and intransigence. In: Proceedings of the European conference on computer vision (ECCV). pp. 532--547 (2018)

\bibitem{FullyCrfs}
Chen, L.C., Papandreou, G., Kokkinos, I., Murphy, K., Yuille, A.L.: Semantic image segmentation with deep convolutional nets and fully connected crfs. arXiv preprint arXiv:1412.7062  (2014)

\bibitem{deeplabv3}
Chen, L.C., Papandreou, G., Kokkinos, I., Murphy, K., Yuille, A.L.: Deeplab: Semantic image segmentation with deep convolutional nets, atrous convolution, and fully connected crfs. IEEE transactions on pattern analysis and machine intelligence  \textbf{40}(4),  834--848 (2017)

\bibitem{simsiam}
Chen, X., He, K.: Exploring simple siamese representation learning. In: Proceedings of the IEEE/CVF conference on computer vision and pattern recognition. pp. 15750--15758 (2021)

\bibitem{maskformer}
Cheng, B., Schwing, A., Kirillov, A.: Per-pixel classification is not all you need for semantic segmentation. Advances in Neural Information Processing Systems  \textbf{34},  17864--17875 (2021)

\bibitem{imagenet}
Deng, J., Dong, W., Socher, R., Li, L.J., Li, K., Fei-Fei, L.: Imagenet: A large-scale hierarchical image database. In: 2009 IEEE conference on computer vision and pattern recognition. pp. 248--255. Ieee (2009)

\bibitem{LWM}
Dhar, P., Singh, R.V., Peng, K.C., Wu, Z., Chellappa, R.: Learning without memorizing. In: Proceedings of the IEEE/CVF conference on computer vision and pattern recognition. pp. 5138--5146 (2019)

\bibitem{vit}
Dosovitskiy, A., Beyer, L., Kolesnikov, A., Weissenborn, D., Zhai, X., Unterthiner, T., Dehghani, M., Minderer, M., Heigold, G., Gelly, S., Uszkoreit, J., Houlsby, N.: An image is worth 16x16 words: Transformers for image recognition at scale. In: International Conference on Learning Representations (2021), \url{https://openreview.net/forum?id=YicbFdNTTy}

\bibitem{PLOP}
Douillard, A., Chen, Y., Dapogny, A., Cord, M.: Plop: Learning without forgetting for continual semantic segmentation. In: Proceedings of the IEEE/CVF conference on computer vision and pattern recognition. pp. 4040--4050 (2021)

\bibitem{PodNet}
Douillard, A., Cord, M., Ollion, C., Robert, T., Valle, E.: Podnet: Pooled outputs distillation for small-tasks incremental learning. In: Computer Vision--ECCV 2020: 16th European Conference, Glasgow, UK, August 23--28, 2020, Proceedings, Part XX 16. pp. 86--102. Springer (2020)

\bibitem{pascalvoc}
Everingham, M., Eslami, S.A., Van~Gool, L., Williams, C.K., Winn, J., Zisserman, A.: The pascal visual object classes challenge: A retrospective. International journal of computer vision  \textbf{111},  98--136 (2015)

\bibitem{catastrophicforgetting}
French, R.M.: Catastrophic forgetting in connectionist networks. Trends in cognitive sciences  \textbf{3}(4),  128--135 (1999)

\bibitem{DAnet}
Fu, J., Liu, J., Tian, H., Li, Y., Bao, Y., Fang, Z., Lu, H.: Dual attention network for scene segmentation. In: Proceedings of the IEEE/CVF conference on computer vision and pattern recognition. pp. 3146--3154 (2019)

\bibitem{byol}
Grill, J.B., Strub, F., Altch{\'e}, F., Tallec, C., Richemond, P., Buchatskaya, E., Doersch, C., Avila~Pires, B., Guo, Z., Gheshlaghi~Azar, M., et~al.: Bootstrap your own latent-a new approach to self-supervised learning. Advances in neural information processing systems  \textbf{33},  21271--21284 (2020)

\bibitem{maskrcnn}
He, K., Gkioxari, G., Doll{\'a}r, P., Girshick, R.: Mask r-cnn. In: Proceedings of the IEEE international conference on computer vision. pp. 2961--2969 (2017)

\bibitem{resnet}
He, K., Zhang, X., Ren, S., Sun, J.: Deep residual learning for image recognition. In: Proceedings of the IEEE conference on computer vision and pattern recognition. pp. 770--778 (2016)

\bibitem{lucir}
Hou, S., Pan, X., Loy, C.C., Wang, Z., Lin, D.: Learning a unified classifier incrementally via rebalancing. In: Proceedings of the IEEE/CVF conference on computer vision and pattern recognition. pp. 831--839 (2019)

\bibitem{Ccnet}
Huang, Z., Wang, X., Huang, L., Huang, C., Wei, Y., Liu, W.: Ccnet: Criss-cross attention for semantic segmentation. In: Proceedings of the IEEE/CVF international conference on computer vision. pp. 603--612 (2019)

\bibitem{EWC}
Kirkpatrick, J., Pascanu, R., Rabinowitz, N., Veness, J., Desjardins, G., Rusu, A.A., Milan, K., Quan, J., Ramalho, T., Grabska-Barwinska, A., et~al.: Overcoming catastrophic forgetting in neural networks. Proceedings of the national academy of sciences  \textbf{114}(13),  3521--3526 (2017)

\bibitem{l2g}
Li, X., Zhou, Y., Wu, T., Socher, R., Xiong, C.: Learn to grow: A continual structure learning framework for overcoming catastrophic forgetting. In: International Conference on Machine Learning. pp. 3925--3934. PMLR (2019)

\bibitem{SPPA}
Lin, Z., Wang, Z., Zhang, Y.: Continual semantic segmentation via structure preserving and projected feature alignment. In: European Conference on Computer Vision. pp. 345--361. Springer (2022)

\bibitem{adaptivenetwork}
Liu, Y., Schiele, B., Sun, Q.: Adaptive aggregation networks for class-incremental learning. In: Proceedings of the IEEE/CVF conference on Computer Vision and Pattern Recognition. pp. 2544--2553 (2021)

\bibitem{FCN}
Long, J., Shelhamer, E., Darrell, T.: Fully convolutional networks for semantic segmentation. In: Proceedings of the IEEE conference on computer vision and pattern recognition. pp. 3431--3440 (2015)

\bibitem{packnet}
Mallya, A., Lazebnik, S.: Packnet: Adding multiple tasks to a single network by iterative pruning. In: Proceedings of the IEEE conference on Computer Vision and Pattern Recognition. pp. 7765--7773 (2018)

\bibitem{RECALL}
Maracani, A., Michieli, U., Toldo, M., Zanuttigh, P.: Recall: Replay-based continual learning in semantic segmentation. In: Proceedings of the IEEE/CVF international conference on computer vision. pp. 7026--7035 (2021)

\bibitem{catastrophicforgetting2}
McCloskey, M., Cohen, N.J.: Catastrophic interference in connectionist networks: The sequential learning problem. In: Psychology of learning and motivation, vol.~24, pp. 109--165. Elsevier (1989)

\bibitem{ILT}
Michieli, U., Zanuttigh, P.: Incremental learning techniques for semantic segmentation. In: Proceedings of the IEEE/CVF international conference on computer vision workshops. pp.~0--0 (2019)

\bibitem{SDR}
Michieli, U., Zanuttigh, P.: Continual semantic segmentation via repulsion-attraction of sparse and disentangled latent representations. In: Proceedings of the IEEE/CVF conference on computer vision and pattern recognition. pp. 1114--1124 (2021)

\bibitem{REMIND}
Phan, M.H., Phung, S.L., Tran-Thanh, L., Bouzerdoum, A., et~al.: Class similarity weighted knowledge distillation for continual semantic segmentation. In: Proceedings of the IEEE/CVF Conference on Computer Vision and Pattern Recognition. pp. 16866--16875 (2022)

\bibitem{sats}
Qiu, Y., Shen, Y., Sun, Z., Zheng, Y., Chang, X., Zheng, W., Wang, R.: Sats: Self-attention transfer for continual semantic segmentation. Pattern Recognition  \textbf{138},  109383 (2023)

\bibitem{icarl}
Rebuffi, S.A., Kolesnikov, A., Sperl, G., Lampert, C.H.: icarl: Incremental classifier and representation learning. In: Proceedings of the IEEE conference on Computer Vision and Pattern Recognition. pp. 2001--2010 (2017)

\bibitem{SGD}
Robbins, H., Monro, S.: A stochastic approximation method. The annals of mathematical statistics pp. 400--407 (1951)

\bibitem{U-net}
Ronneberger, O., Fischer, P., Brox, T.: U-net: Convolutional networks for biomedical image segmentation. In: Medical Image Computing and Computer-Assisted Intervention--MICCAI 2015: 18th International Conference, Munich, Germany, October 5-9, 2015, Proceedings, Part III 18. pp. 234--241. Springer (2015)

\bibitem{HP}
Seong, H.S., Moon, W., Lee, S., Heo, J.P.: Leveraging hidden positives for unsupervised semantic segmentation. In: Proceedings of the IEEE/CVF Conference on Computer Vision and Pattern Recognition. pp. 19540--19549 (2023)

\bibitem{Incrementer}
Shang, C., Li, H., Meng, F., Wu, Q., Qiu, H., Wang, L.: Incrementer: Transformer for class-incremental semantic segmentation with knowledge distillation focusing on old class. In: Proceedings of the IEEE/CVF Conference on Computer Vision and Pattern Recognition. pp. 7214--7224 (2023)

\bibitem{ganreplay}
Shin, H., Lee, J.K., Kim, J., Kim, J.: Continual learning with deep generative replay. Advances in neural information processing systems  \textbf{30} (2017)

\bibitem{robotics}
Siam, M., Elkerdawy, S., Jagersand, M., Yogamani, S.: Deep semantic segmentation for automated driving: Taxonomy, roadmap and challenges. In: 2017 IEEE 20th international conference on intelligent transportation systems (ITSC). pp.~1--8. IEEE (2017)

\bibitem{Segmenter}
Strudel, R., Garcia, R., Laptev, I., Schmid, C.: Segmenter: Transformer for semantic segmentation. In: Proceedings of the IEEE/CVF international conference on computer vision. pp. 7262--7272 (2021)

\bibitem{gcr}
Tiwari, R., Killamsetty, K., Iyer, R., Shenoy, P.: Gcr: Gradient coreset based replay buffer selection for continual learning. In: Proceedings of the IEEE/CVF Conference on Computer Vision and Pattern Recognition. pp. 99--108 (2022)

\bibitem{ConSeg}
Wang, W., Zhou, T., Yu, F., Dai, J., Konukoglu, E., Van~Gool, L.: Exploring cross-image pixel contrast for semantic segmentation. In: Proceedings of the IEEE/CVF International Conference on Computer Vision. pp. 7303--7313 (2021)

\bibitem{segformer}
Xie, E., Wang, W., Yu, Z., Anandkumar, A., Alvarez, J.M., Luo, P.: Segformer: Simple and efficient design for semantic segmentation with transformers. Advances in Neural Information Processing Systems  \textbf{34},  12077--12090 (2021)

\bibitem{der}
Yan, S., Xie, J., He, X.: Der: Dynamically expandable representation for class incremental learning. In: Proceedings of the IEEE/CVF Conference on Computer Vision and Pattern Recognition. pp. 3014--3023 (2021)

\bibitem{uac}
Yang, G., Fini, E., Xu, D., Rota, P., Ding, M., Nabi, M., Alameda-Pineda, X., Ricci, E.: Uncertainty-aware contrastive distillation for incremental semantic segmentation. IEEE Transactions on Pattern Analysis and Machine Intelligence  \textbf{45}(2),  2567--2581 (2022)

\bibitem{LGKD}
Yang, Z., Li, R., Ling, E., Zhang, C., Wang, Y., Huang, D., Ma, K.T., Hur, M., Lin, G.: Label-guided knowledge distillation for continual semantic segmentation on 2d images and 3d point clouds. In: Proceedings of the IEEE/CVF International Conference on Computer Vision. pp. 18601--18612 (2023)

\bibitem{csi}
Zenke, F., Poole, B., Ganguli, S.: Continual learning through synaptic intelligence. In: International conference on machine learning. pp. 3987--3995. PMLR (2017)

\bibitem{RC}
Zhang, C.B., Xiao, J.W., Liu, X., Chen, Y.C., Cheng, M.M.: Representation compensation networks for continual semantic segmentation. In: Proceedings of the IEEE/CVF Conference on Computer Vision and Pattern Recognition. pp. 7053--7064 (2022)

\bibitem{RBC}
Zhao, H., Yang, F., Fu, X., Li, X.: Rbc: Rectifying the biased context in continual semantic segmentation. In: European Conference on Computer Vision. pp. 55--72. Springer (2022)

\bibitem{ade20k}
Zhou, B., Zhao, H., Puig, X., Fidler, S., Barriuso, A., Torralba, A.: Scene parsing through ade20k dataset. In: Proceedings of the IEEE conference on computer vision and pattern recognition. pp. 633--641 (2017)

\end{thebibliography}
\end{document}